\documentclass[preprint,review,12pt]{elsarticle}

\usepackage[left=2.5cm, right=2.5cm, top=2.5cm, bottom=2.5cm]{geometry}
\usepackage{amssymb}
\usepackage{graphicx}%
\usepackage{multirow}%
\usepackage{amssymb,amsfonts}%
\usepackage{amsmath}%
\usepackage{amsthm}%
\usepackage{mathrsfs}%
\usepackage[title]{appendix}%
\usepackage{xcolor}%
\usepackage{textcomp}%
\usepackage{manyfoot}%
\usepackage{booktabs}%
\usepackage{algorithm}%
\usepackage{algpseudocode}%
\usepackage{listings}%
\usepackage{color,soul}

\usepackage{float}
\usepackage{caption}
\usepackage{subcaption}
\usepackage{tabularx}
\usepackage{amssymb}
\usepackage{pifont}
\usepackage{hyperref}
\usepackage{setspace}
\usepackage{placeins}
\journal{Elsevier}
\begin{document}

\begin{frontmatter}



\title{CARLA: Self-supervised Contrastive Representation Learning for Time Series Anomaly Detection}

\cortext[cor1]{Corresponding author}
\author[inst1]{Zahra Zamanzadeh~Darban\corref{cor1}}
\ead{zahra.zamanzadeh@monash.edu}

\author[inst1]{Geoffrey I.~Webb}
\author[inst2]{Shirui Pan}
\author[inst3]{Charu C.~Aggarwal}
\author[inst1]{Mahsa Salehi}

\affiliation[inst1]{organization={Monash University},
            city={Melbourne}, 
            state={Victoria},
            country={Australia}}
            
\affiliation[inst2]{organization={Griffith University},
    city={Gold Coast},
    state={Queensland},
    country={Australia}}

\affiliation[inst3]{organization={IBM T. J. Watson Research Center},
    city={Yorktown Heights},
    state={NY},
    country={USA}}

\begin{abstract}
One main challenge in time series anomaly detection (TSAD) is the lack of labelled data in many real-life scenarios. Most of the existing anomaly detection methods focus on learning the normal behaviour of unlabelled time series in an unsupervised manner. The normal boundary is often defined tightly, resulting in slight deviations being classified as anomalies, consequently leading to a high false positive rate and a limited ability to generalise normal patterns. To address this, we introduce a novel end-to-end self-supervised ContrAstive Representation Learning approach for time series Anomaly detection (CARLA). While existing contrastive learning methods assume that augmented time series windows are positive samples and temporally distant windows are negative samples, we argue that these assumptions are limited as augmentation of time series can transform them to negative samples, and a temporally distant window can represent a positive sample. Existing approaches to contrastive learning for time series have directly copied methods developed for image analysis. We argue that these methods do not transfer well. Instead, our contrastive approach leverages existing generic knowledge about time series anomalies and injects various types of anomalies as negative samples. Therefore, CARLA not only learns normal behaviour but also learns deviations indicating anomalies. It creates similar representations for temporally close windows and distinct ones for anomalies. Additionally, it leverages the information about representations' neighbours through a self-supervised approach to classify windows based on their nearest/furthest neighbours to further enhance the performance of anomaly detection. In extensive tests on seven major real-world TSAD datasets, CARLA shows superior performance (F1 and AU-PR) over state-of-the-art self-supervised, semi-supervised, and unsupervised TSAD methods for univariate time series and multivariate time series. Our research highlights the immense potential of contrastive representation learning in advancing the TSAD field, thus paving the way for novel applications and in-depth exploration.
\end{abstract}


% \begin{highlights}
% \item CARLA is a novel contrastive learning-based time series anomaly detection framework.

% \item CARLA uses anomaly injection to create negative examples for contrastive learning.

% \item CARLA's effectiveness is validated through an extensive and robust evaluation approach.

% \end{highlights}


\end{frontmatter}


\section{Introduction} \label{sec1}

In many modern applications, data analysis is required to identify and remove anomalies (a.k.a. outliers) to ensure system reliability. Several machine learning algorithms are well-suited for detecting these outliers~\cite{domingues2018comparative}. In time series data, anomalies can result from different factors, including equipment failure, sensor malfunction, human error, and human intervention. Detecting anomalies in time series data has numerous real-world uses, including monitoring equipment for malfunctions, detecting unusual patterns in IoT sensor data, enhancing the reliability of computer programs and cloud systems, observing patients' health metrics, and pinpointing cyber threats.
Time series anomaly detection (TSAD) has been the subject of decades of intensive research, with numerous approaches proposed to address the challenge of spotting rare and unexpected events in complex and noisy data. Statistical methods have been developed to monitor and identify abnormal behaviour~\cite{domingues2018comparative}. Recent advancements in deep learning techniques have effectively tackled various anomaly detection problems~\cite{pang2021deep}. Specifically, for time series with complex nonlinear temporal dynamics, deep learning methods have demonstrated remarkable performance~\cite{schmidl2022anomaly}.

\begin{figure}[t]
    \centering
    \begin{subfigure}[b]{0.25\columnwidth}
        \centering
        \includegraphics[width=\textwidth]{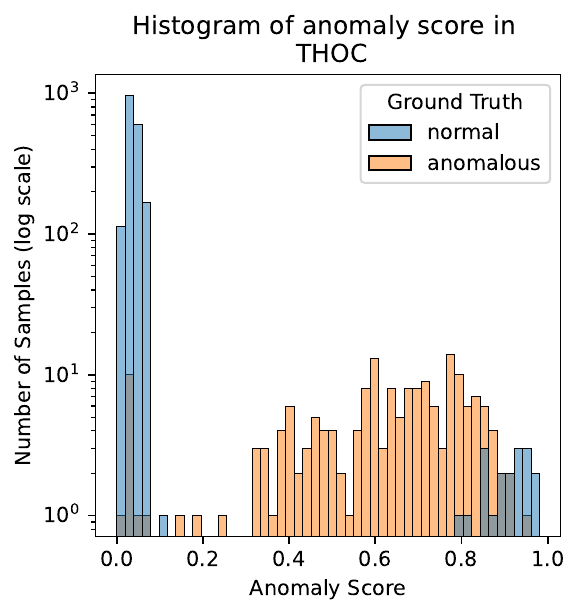}
        \caption{}
        \label{fig:thoc-dist}
    \end{subfigure}
    \begin{subfigure}[b]{0.25\columnwidth}
        \centering
        \includegraphics[width=\textwidth]{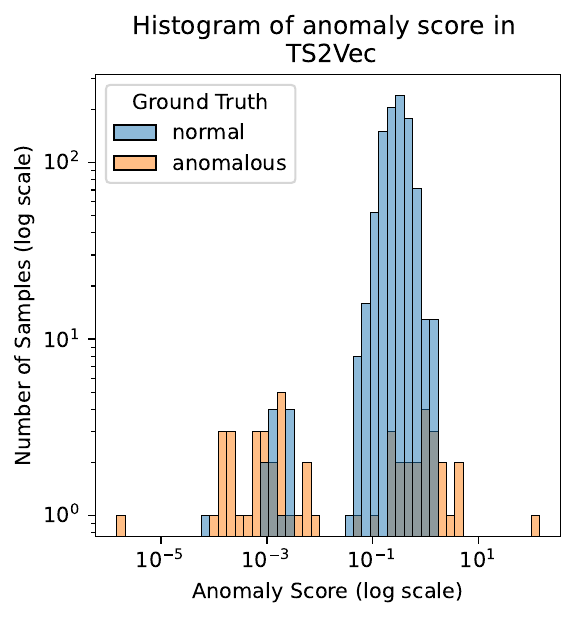}
        \caption{}
        \label{fig:ts2vec-dist}
    \end{subfigure}
    \begin{subfigure}[b]{0.25\columnwidth}
        \centering
        \includegraphics[width=\textwidth]{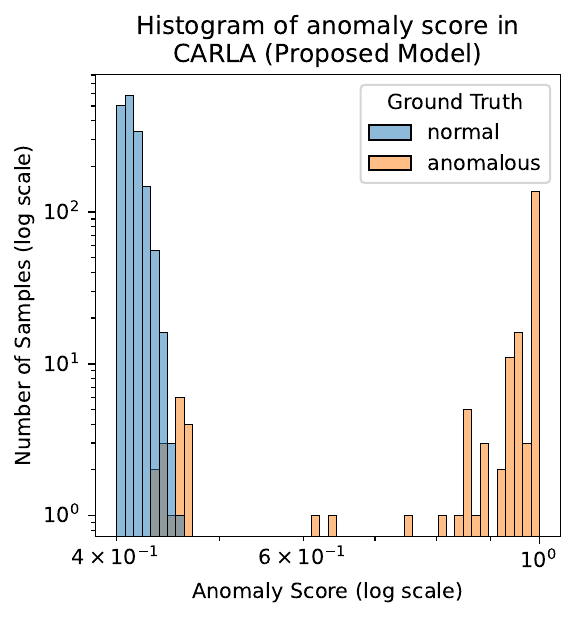}
        \caption{}
        \label{fig:carla-dist}
    \end{subfigure}
    \caption{Histograms of the distribution of anomaly scores produced by (\subref{fig:thoc-dist}) THOC~\cite{shen2020timeseries}, (\subref{fig:ts2vec-dist}) and TS2Vec~\cite{yue2022ts2vec}
    and (\subref{fig:carla-dist}) CARLA models
    using M-6 dataset of the MSL benchmark~\cite{hundman2018detecting}.}
    \label{fig:discriminate}
\end{figure}

Most of these models focus on learning the normal behaviour of data from an unlabelled dataset and, therefore, can potentially predict samples that deviate from the normal behaviour as anomalies. The lack of labelled data in real-world scenarios makes it difficult for models to learn the difference between normal and anomalous behaviours. The normal boundary is often tightly defined, which can result in slight deviations being classified as anomalies. Models that cannot discriminate between normal and anomaly classes may predict normal samples as anomalous, leading to high false positives. Figure~\ref{fig:thoc-dist} shows an example of a histogram of anomaly scores of an existing unsupervised TSAD method called THOC~\cite{shen2020timeseries} on a benchmark dataset (M-6 test set from the MSL dataset).
Frequent false positives often occur when normal data are assigned high anomaly scores. In instances where the representation of a segment of the normal data closely resembles that of anomalous data, there can be an elevated anomaly score attributed to these normal samples, consequently leading to an increase in false positives. This issue can be seen on the right side of Figure~\ref{fig:thoc-dist}, where many normal samples are assigned high anomaly scores.

An alternative approach is to leverage self-supervised contrastive representation learning. Contrastive learning entails training a model to differentiate between pairs of similar and dissimilar samples in a dataset. Contrastive representation learning has been successful in image~\cite{chen2020simple} and natural language processing~\cite{logeswaran2018}, and its potential in TSAD has been explored in recent years. In the context of an anomaly detection task where the majority of the data is normal, contrastive loss functions can increase the distance between normal samples and their corresponding negative (anomalous) samples. Simultaneously, they decrease the distance between normal samples and their corresponding positive (normal) samples. By doing so, the distinction between normal and anomalous data is made clearer and more pronounced, which helps to identify a more accurate boundary for the normal samples. However, existing contrastive learning methods for TSAD mainly assume that augmented time series windows are positive samples and temporally distant windows are negative samples. We argue that these assumptions carry the risk that augmentation of time series can transform them to negative samples, and a temporally distant window can represent a positive sample, leading to ineffective anomaly detection performance. Figure~\ref{fig:ts2vec-dist} shows anomaly scores of a contrastive learning TSAD method called TS2Vec~\cite{yue2022ts2vec}. As demonstrated, the anomaly scores of both normal and anomalous data are blended together, resulting in the inefficacy of anomaly detection. In fact, in situations where normal and anomalous data are intermingled within the representation space, achieving an enhanced anomaly detection rate—as measured by metrics such as the F1 score or the area under the precision-recall curve (AU-PR)—is accompanied by an increase in the false positive rate.

We propose a novel two-stage framework called CARLA, designed specifically to enhance time series anomaly detection. Our novel approach addresses the lack of labelled data through a contrastive approach, which leverages existing generic knowledge about different types of time series anomalies~\cite{darban2022deep} in the first stage (pretext). We inject various types of anomalies, which facilitates the learning representations of normal behaviour. This is achieved by encouraging the learning of similar representations for windows that are temporally closed windows while ensuring dissimilar representations for windows and their corresponding injected anomalous windows.

Additionally, to ensure that the representation of existing real anomalous windows (for which we do not have labels) is different from normal representations, we employ a self-supervised approach to classify normal/anomalous representations of windows based on their nearest/furthest neighbours in the representation space in the second stage (self-supervised classification). By making the normal representations more discriminative, we enhance the anomaly detection performance in our proposed model.
Specifically, the main contributions of this paper can be summarised as follows:
\begin{itemize}
\item We propose a novel contrastive representation learning model to detect anomalies in time series, which delivers top-tier outcomes across a range of real-world benchmark datasets, encompassing both univariate time series (UTS) and multivariate time series (MTS). Addressing the challenge of lack of labelled data, our model learns to effectively discriminate normal patterns from anomalous ones in the feature representation space (see Figure~\ref{fig:carla-dist}). 
CARLA's implementation is publicly available on Github: \url{https://github.com/zamanzadeh/CARLA}.  

\item We propose an effective contrastive method for TSAD to learn feature representations for a pretext task by leveraging existing generic knowledge about time series anomalies (see Figure~\ref{fig:tsne-pretext}).
\item We propose a self-supervised classification method that leverages the representations learned in the pretext stage to classify time series windows. Our goal is to classify each sample by utilising its neighbours in the representation space, which were learned during the pretext stage (see Figure~\ref{fig:tsne-class}).
\item Our comprehensive analysis across seven real-world benchmark datasets reveals the superior performance of CARLA over a range of ten SOTA unsupervised, semi-supervised, and self-supervised contrastive learning models. CARLA's consistent balance between False Positive Rate (FPR) and Area Under the Precision-Recall Curve (AU-PR) throughout various MTS and UTS datasets underscores its precision and reliability. This balance ensures that CARLA provides reliable and precise alerts, which are crucial for many real-world applications.

\end{itemize}

\section{Related Work}\label{sec2}
In this section, we concentrate on three areas: deep learning methods in TSAD, unsupervised representation of time series, and the contrastive representation learning technique.

\subsection{Time Series Anomaly Detection}
The detection of anomalies within time series data has been the subject of extensive research, using an array of techniques from statistical methods to classical machine learning and, more recently, deep learning models~\cite{schmidl2022anomaly}. Established statistical techniques such as moving averages like the ARIMA model~\cite{box1970distribution} have seen widespread application. Machine learning techniques, including clustering algorithms and density-based approaches, alongside algorithms similar to decision trees~\cite{liu2008isolation} have also been leveraged.

\textbf{Deep Learning methods in TSAD: } 
In recent years, deep learning has proven highly effective due to its ability to autonomously extract features~\cite{pang2021deep}. The focus within TSAD is largely on unsupervised~\cite{hundman2018detecting}, semi-supervised~\cite{park2018multimodal}, and recently self-supervised~\cite{yue2022ts2vec} approaches, addressing the issue of scarce labelled data. Techniques such as OmniAnomaly~\cite{su2019robust} prove to be particularly useful in situations where anomaly labels are unavailable, while semi-supervised methods make efficient use of the labels that are available.

Deep learning techniques, encompassing autoencoders~\cite{yao2023regularizing}, variational autoencoders (VAEs)~\cite{park2018multimodal}, RNNs~\cite{su2019robust}, LSTM networks~\cite{hundman2018detecting}, GANs~\cite{li2019mad}, and Transformers~\cite{Tuli2022TranADDT,xu2021anomaly} have shown potential in TSAD, particularly for high-dimensional or non-linear data.
The preference for deep models like LSTM-VAE~\cite{park2018multimodal}, DITAN~\cite{giannoulis2023ditan} and THOC~\cite{shen2020timeseries} is because they excel at minimising forecasting errors while capturing time series data's temporal dependencies.

In summary, semi-supervised methods, such as LSTM-VAE~\cite{park2018multimodal} excel when labels are readily available. On the other hand, unsupervised methods, such as OmniAnomaly~\cite{su2019robust} and AnomalyTransformer~\cite{xu2021anomaly}, become more suitable when obtaining anomaly labels is a challenge. These unsupervised deep learning methods are preferred due to their capability to learn robust representations without requiring labelled data. The advent of self-supervised learning methods has further improved generalisation in unsupervised anomaly detection~\cite{zhao2020multivariate}.

\subsection{Unsupervised Time Series Representation}
The demonstration of substantial performance by unsupervised representation learning across a wide range of fields, such as computer vision~\cite{chen2020simple, milbich2020unsupervised}, natural language processing~\cite{logeswaran2018}, and speech recognition~\cite{xu2021}, has made it a desirable technique. In the realm of time series, methods like TKAE~\cite{bianchi2019learning} and TST~\cite{zerveas2021} have been suggested. While these methods have made significant contributions, some of them face challenges with scalability for exceptionally long time series or encounter difficulties in modelling complex time series. To overcome these limitations, methods such as TNC~\cite{tonekaboni2021} and T-Loss~\cite{franceschi2019} have been proposed. They leverage time-based negative sampling, triplet loss, and local smoothness of signals for the purpose of learning scalable MTS representations. Despite their merits, these methods often limit their universality by learning representations of particular semantic tiers, relying on significant assumptions regarding invariance during transformations. Recent studies, such as RoSAS \cite{xu2023rosas}, have outlier detection in datasets beyond time series by utilising triplet loss.

\subsection{Contrastive Representation Learning}
Contrastive representation learning creates an embedding space where similar samples are close and dissimilar ones are distant, applicable in domains like natural language processing, computer vision, and time series anomaly detection~\cite{le2020contrastive}. Traditional models like InfoNCE loss~\cite{oord2018representation} and SimCLR~\cite{chen2020simple} use positive-negative pairs to optimise this space, demonstrating strong performance and paving the way for advanced techniques.
% Siamese architectures such as SimSiam~\cite{chen2021exploring} represent a significant advancement by eliminating the need for negative samples, focusing solely on positive pairs. These models simplify the architecture while maintaining competitive performance.
In TSAD, contrastive learning is crucial for recognising patterns. TS2Vec~\cite{yue2022ts2vec} uses this approach hierarchically for multi-level semantic representation, while DCdetector~\cite{yang2023dcdetector} introduces a dual attention asymmetric design with pure contrastive loss, enabling permutation invariant representations.

\section{CARLA}\label{sec3}
\textbf{Problem definition:}
Given a time series $\mathcal{D}$ which is partitioned into $m$ overlapping time series windows $\{ w_1, .., w_i, .., w_m \}$ with stride 1 where $w_i=\{x_1,.., x_i, .., x_{WS}\}$, $WS$ is time series window size, $x_i \in \mathbb{R}^{Dim}$ and $Dim$ is the dimension of time series, the goal is to detect anomalies in time series windows.

\begin{figure*}[t]
    \centering
    \includegraphics[width=1.0\linewidth]{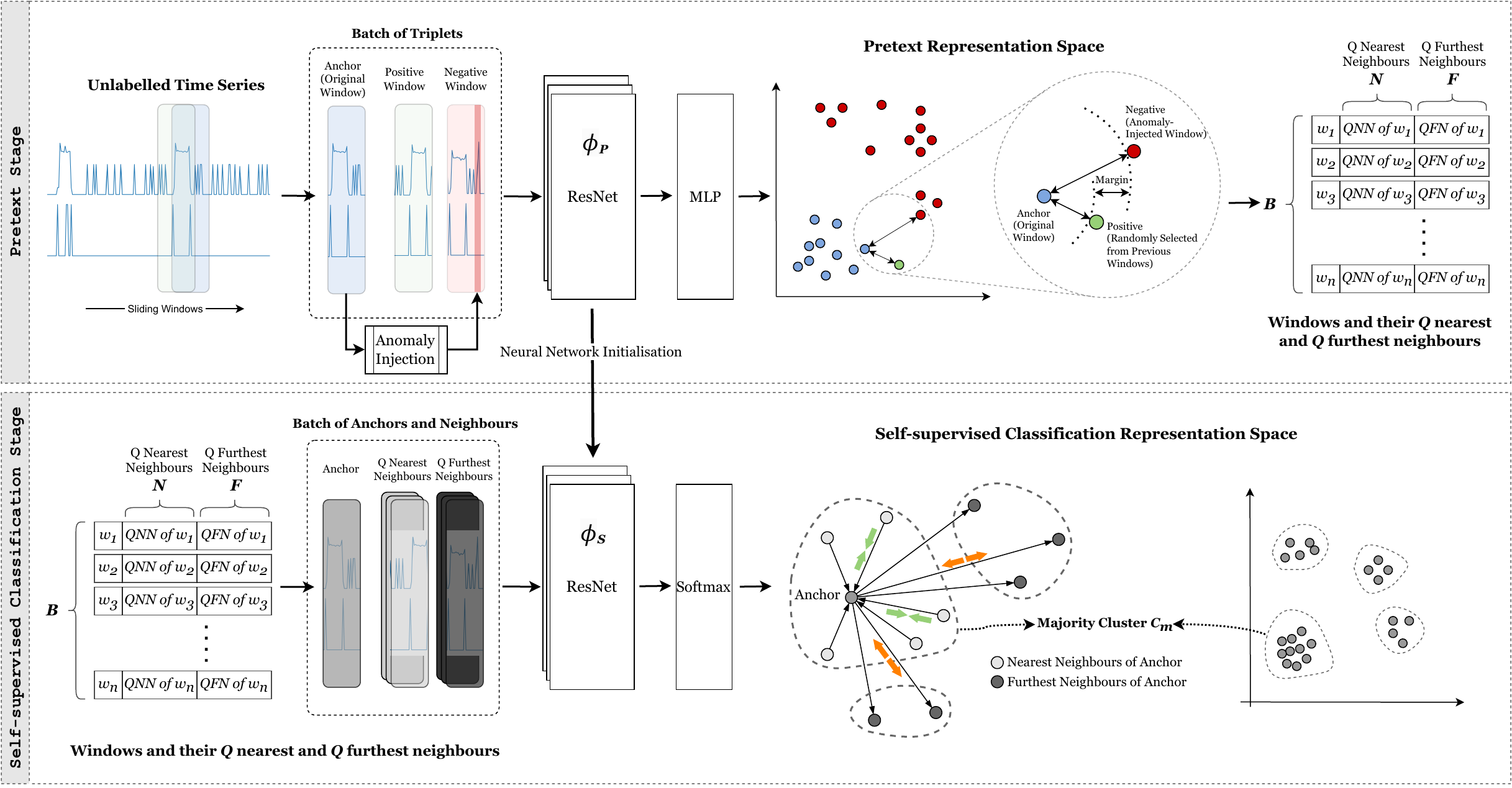}
    \caption {The end-to-end pipeline of CARLA consists of two main stages: the Pretext Stage and the Self-supervised Classification Stage. In the Pretext Stage, anomaly injection techniques are used for self-supervised learning. The Self-supervised Classification Stage integrates the learned representations for a contrastive approach that maximises the similarity between anchor samples and their nearest neighbours while minimising the similarity between anchor samples and their furthest neighbours. The output is a trained model and the majority class, enabling inference for anomaly detection.}
    \label{fig:method}
\end{figure*}

CARLA (A Self-supervised \textbf{C}ontr\textbf{A}stive \textbf{R}epresentation \textbf{L}earning Approach for time series \textbf{A}nomaly detection) is built on several key components, each of which plays a critical role in achieving effective representation learning as illustrated in Figure~\ref{fig:method}. CARLA consists of two main stages: the Pretext Stage and the Self-supervised Classification Stage. 

Initially, in the Pretext Stage (section \ref{subsec3.1}), it employs anomaly injection to learn similar representations for temporally proximate windows and distinct representations for windows and their equivalent anomalous windows (section \ref{subsec3.0}). These injected anomalies include point anomalies, such as sudden spikes, and subsequence anomalies, like unexpected pattern shifts. This technique not only aids in training the model to recognise deviations from the norm but also strengthens its ability to generalise across various types of anomalies. At the end of the Pretext Stage, we establish a prior by finding the nearest and furthest neighbours for each window representation for the next stage. The Self-supervised Classification Stage (section \ref{subsec3.2}) then classifies these window representations as normal or anomalous based on the proximity of their neighbours in the representation space (section \ref{subsec3.3}). This classification aims to group similar time series windows together while distinctly separating them from dissimilar ones. The effectiveness of this stage is pivotal in accurately categorising time series windows, reinforcing CARLA's capability to differentiate between normal and anomalous patterns. The comprehensive end-to-end pipeline of CARLA, including these stages, is illustrated in Figure~\ref{fig:method}.
The following sections present the ideas underlying our approach. 

\subsection{Anomaly Injection} \label{subsec3.0}

The technique of anomaly injection, a powerful data augmentation strategy for time series, facilitates the application of self-supervised learning in the absence of ground-truth labels. This technique has recently been applied in TSAD, notably in COUTA~\cite{xu2024calibrated}. While the augmentation methods we employ are not designed to represent every conceivable anomaly type~\cite{xu2021anomaly} — a goal that would be unattainable — they amalgamate various robust and generic heuristics to effectively identify prevalent out-of-distribution instances.

\subsubsection{Anomaly Injection Steps}

During the training phase, each window is manipulated by randomly choosing instances within a given window $w_i$. Two primary categories of anomaly injection models—point anomalies and subsequence anomalies—are adopted to inject anomalies to a time series window $w_i$. In a multivariate time series context, a random start time and a subset of dimension(s) $d$ are selected for the injection of a point or subsequence anomalies. Note in the context of multivariate time series, anomalies are not always present across all dimensions, prompting us to randomly select a subset of dimensions for the induced anomalies ($d < \lceil Dim/10 \rceil $). The injected anomaly portion for each dimension varied from 1 data point to 90\% of the window length. 

This approach fosters the creation of a more diverse set of anomalies, enhancing our model's capability of detecting anomalies that exist in multiple dimensions. This approach strengthens our model's effectiveness in discriminating between normal and anomalous representations in the Pretext Stage. It is important to underscore that the anomaly injection strategies employed in our model's evaluation were consistently applied across all benchmarks to ensure a fair and impartial comparison of the model's performance across various datasets. The step-by-step process for our anomaly injection approach is detailed in Algorithm~\ref{algo:carlainject}. This comprehensive algorithm encapsulates the methodology of point and subsequence anomaly injection, providing a clear understanding of our process and its implementation.

\begin{algorithm}[t]
\scriptsize
  \caption{\texttt{InjectAnomaly}($w$)}
  \label{algo:carlainject}
  \begin{algorithmic}[1]
  \renewcommand{\algorithmicrequire}{\textbf{Input:}}
  \renewcommand{\algorithmicensure}{\textbf{Output:}}

\Require Time series window $w$
\Ensure Anomaly injected time series window $w'$
\State $types \gets \{\texttt{Seasonal}, \texttt{Trend}, \texttt{Global}, \texttt{Contextual}, \texttt{Shapelet}\} $
\State $Dim \gets \text{random subset of dimensions of } w ~\text{(for UTS, } Dim=\{1\}\text{)}$
\State $w' \gets \text{copy of}~w$
\State $s, e \gets \text{random start and end points from } (0,size(w)] \text{, where } e>s $
\For {each dimension $d$ in $Dim$}
    \State $anomaly \gets \text{randomly choose anomaly type from } types $
    \If {$anomaly = \texttt{Global}$}
        \State $\mu_{w}, \sigma_{w} \gets \text{mean and std of } w \text{ in dimension } d$
        \State $g \gets \text{random number form }[3, 5]$ \hfill $\triangleright$ \text{coefficient for \texttt{Global} anomaly}
        \State $w'(s) = \text{random value from } \{\mu_{w} + g.\sigma_{w}, \mu_{w} - g.\sigma_{w}\}$
    \ElsIf {$anomaly = \texttt{Contextual}$}
        \State $\mu_{se}, \sigma_{se} \gets \text{mean and std of subsequence } w(t | t \in [s, e]) \text{ in dimension } d$
        \State $x \gets \text{random number form }[3, 5]$ \hfill $\triangleright$ \text{coefficient for \texttt{Contextual} anomaly}
        \State $w'(s) = \text{random value from } \{\mu_{se} + x.\sigma_{se}, \mu_{se} - x.\sigma_{se}\}$
    \ElsIf {$anomaly = \texttt{Seasonal}$}
        \State $f \gets \text{random number from } \{\frac{1}{3}, \frac{1}{2}, 2, 3\}$ \hfill $\triangleright$ \text{frequency coefficient for \texttt{Seasonal} anomaly}
        \State $w'(t) = \begin{cases} 
        w(s + \left(\left\lfloor(t - s) \cdot f\right\rfloor \mod n\right)) & \text{if } s \leq t < e \text{ and } f > 1 \\
        w(s + \left\lfloor(t - s) \cdot f\right\rfloor) & \text{if } s \leq t < e \text{ and } 0 < f < 1 \\
        w(t) & \text{otherwise} 
        \end{cases}$
    \ElsIf {$anomaly = \texttt{Trend}$}
        \State $b \gets \text{random number form }[3, 5]$ \hfill $\triangleright$ \text{coefficient for \texttt{Trend} anomaly}
        \State $w'(t) = \begin{cases} 
        w(t)+b.\sigma_{w} & \text{if } s \leq t \leq e \\
        w(t) & \text{if } t < s \text{ or } t > e
        \end{cases}$ 
    \ElsIf {$anomaly = \texttt{Shapelet}$}
        \State $w'(t) = \begin{cases} 
        w(s) & \text{if } s \leq t \leq e \\
        w(t) & \text{if } t < s \text{ or } t > e 
        \end{cases}$
    \EndIf
\EndFor
\State \textbf{Return} $w'$
\end{algorithmic}
\end{algorithm}

\subsubsection{Point Anomalies}
Our methodology incorporates the injection of single-point anomalies (a spike) at randomly chosen instances within a given window \(w_i\). We employ two distinct types of point anomalies, namely, \texttt{Global} (Figure~\ref{fig:anomlay_Global}) and \texttt{Contextual} (Figure~\ref{fig:anomlay_Contextual})~\cite{xu2021anomaly}. 

\subsubsection{Subsequence Anomalies}
\begin{figure}[t]
     \centering
    \begin{subfigure}[b]{0.22\textwidth}
        \centering
        \includegraphics[trim={0 0 0 0},clip,width=\textwidth]{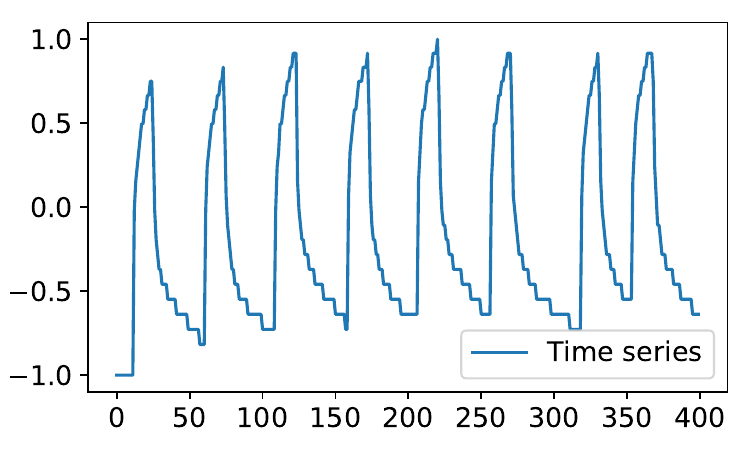}
        \caption{Original Window}\vspace*{10pt}
        \label{fig:original_window}
    \end{subfigure}
    \begin{subfigure}[b]{0.22\textwidth}
        \centering
        \includegraphics[trim={0 0 0 0},clip,width=\textwidth]{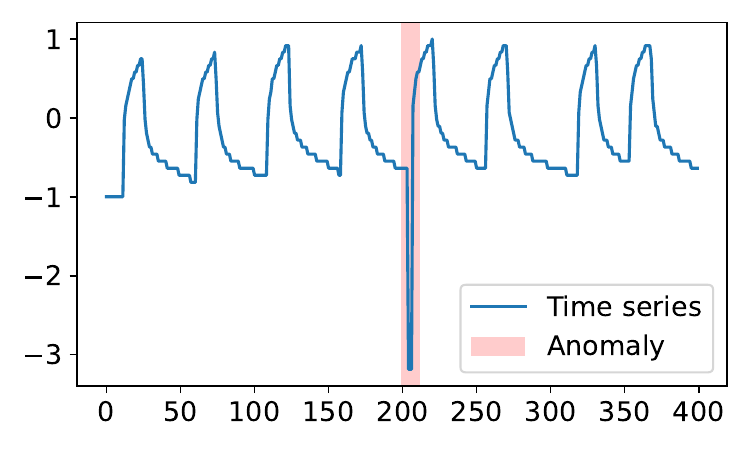}
        \caption{Global Anomaly}\vspace*{10pt}
        \label{fig:anomlay_Global}
    \end{subfigure}
    \begin{subfigure}[b]{0.22\textwidth}
        \centering
        \includegraphics[trim={0 0 0 0},clip,width=\textwidth]{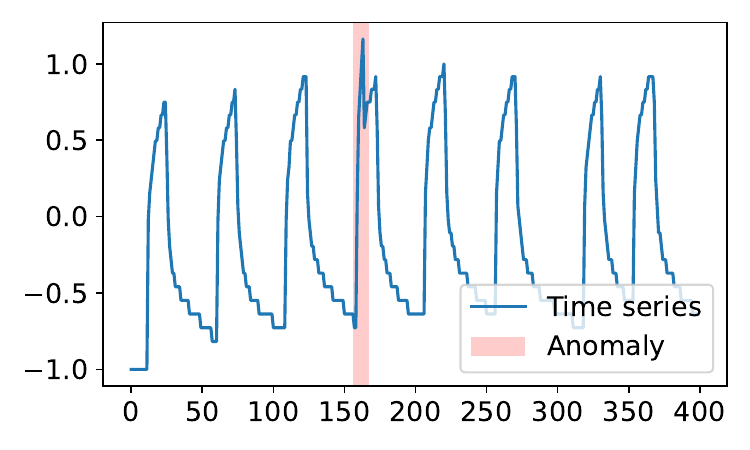}
         \caption{Contextual Anomaly}\vspace*{10pt}
        \label{fig:anomlay_Contextual}
    \end{subfigure}\\
    \begin{subfigure}[b]{0.22\textwidth}
        \centering
        \includegraphics[trim={0 0 0 0},clip,width=\textwidth]{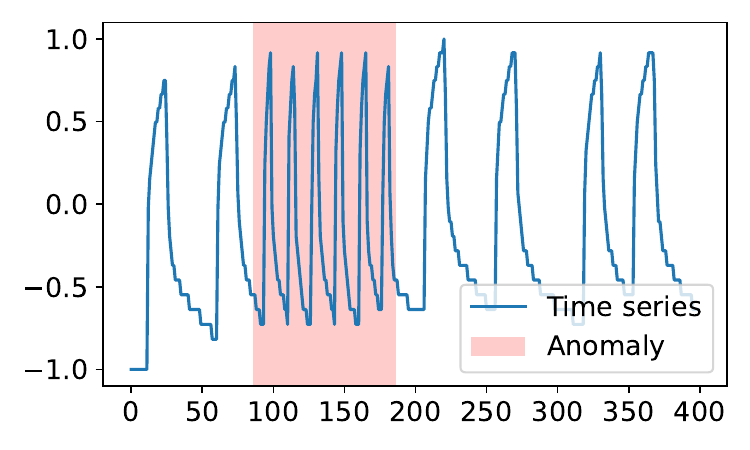}
        \caption{Seasonal Anomaly}
        \label{fig:anomlay_Seasonal}
    \end{subfigure}
    \begin{subfigure}[b]{0.22\textwidth}
        \centering
        \includegraphics[trim={0 0 0 0},clip,width=\textwidth]{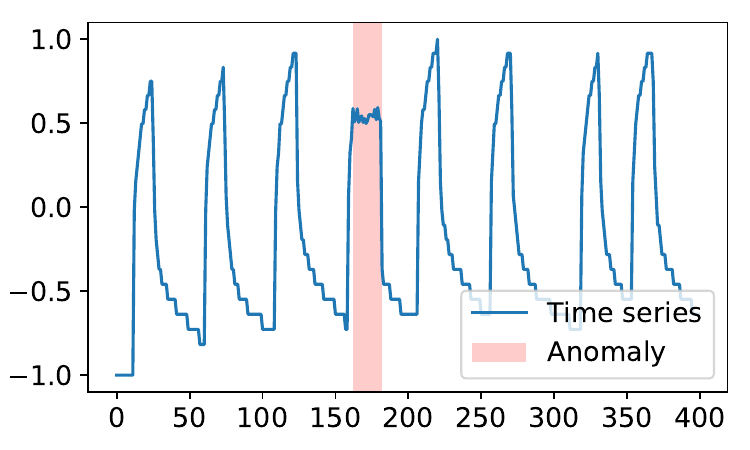}
         \caption{Shapelet Anomaly}
        \label{fig:anomlay_Shapelet}
    \end{subfigure}
    \begin{subfigure}[b]{0.22\textwidth}
        \centering
        \includegraphics[trim={0 0 0 0},clip,width=\textwidth]{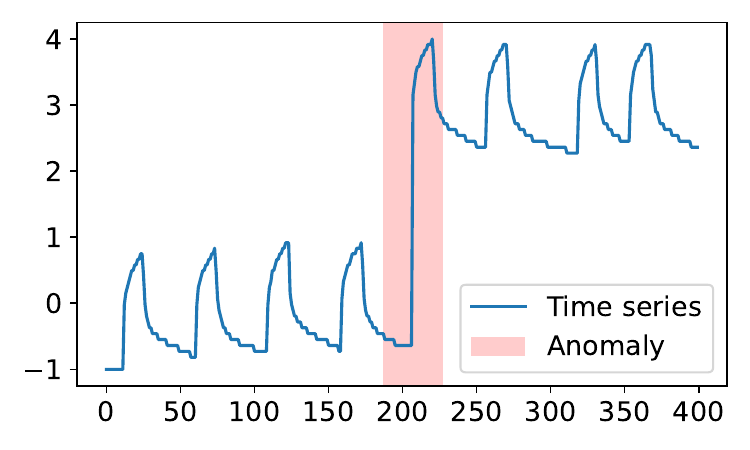}
         \caption{Trend Anomaly}
        \label{fig:anomlay_Trend}
    \end{subfigure}
    
    \caption {Different types of synthetic anomaly injection used in CARLA. The figure presents the effect of synthetic anomaly injections into a randomly selected window of size 400 from the first dimension of entity E-2 in the MSL dataset~\citep{hundman2018detecting}. The (\subref{fig:original_window}) represents the original time series window, while the remaining five demonstrate the same window but with different types of anomalies injected. The anomalies are categorised into (\subref{fig:anomlay_Global}) \texttt{Global}, (\subref{fig:anomlay_Contextual}) \texttt{Contextual}, (\subref{fig:anomlay_Seasonal}) \texttt{Seasonal}, (\subref{fig:anomlay_Shapelet}) \texttt{Shapelet}, and (\subref{fig:anomlay_Trend}) \texttt{Trend}. In each of the five plots with anomalies, the anomalous points or subsequences are accentuated with a red colour.}
    % The synthetic anomalies were introduced using one or more randomly selected anomaly injection methods, applied sequentially across one or more dimensions of the multivariate time series.}
    \label{fig:injection}
\end{figure}

The technique of subsequence anomaly injection, motivated by the successful implementation of the Outlier Exposure approach~\cite{hendrycks2018deep}, enhances anomaly detection in time series data by generating contextual out-of-distribution examples. It involves the introduction of a subsequence anomaly within a time series window \(w_i\), represented as \(w(t | t \in [s, e])\), where \(s\) and \(e\) denotes the anomaly's starting and end time points. We explore three primary types of subsequence (a.k.a. pattern) anomalies: \texttt{Seasonal} (Figure~\ref{fig:anomlay_Seasonal}), \texttt{Shapelet} (Figure~\ref{fig:anomlay_Shapelet}), and \texttt{Trend} (Figure~\ref{fig:anomlay_Trend})~\cite{xu2021anomaly}. 

\subsection{Pretext Stage} \label{subsec3.1}
The Pretext Stage of CARLA consists of two parts: Part One is a contrastive representation learning step that uses a ResNet architecture (which has shown effectiveness in times series classification tasks~\cite{ismail2019deep}) to learn representations for time series windows, and Part Two is a post-processing step that uses the learned representations from Part One to identify semantically meaningful nearest and furthest neighbours for each learned window representation. Algorithm \ref{algo:carlapretext} shows the Pretext steps.
\begin{algorithm}[t]
\scriptsize
  \caption{\texttt{PretextCARLA}($\mathcal{D}$, $Q$, $\alpha$)}
  \label{algo:carlapretext}
  \begin{algorithmic}[1]
  \renewcommand{\algorithmicrequire}{\textbf{Input:}}
  \renewcommand{\algorithmicensure}{\textbf{Output:}}
    \Require Sequential time series windows $\mathcal{D} = \{w_1, w_2, ..., w_m\}$, number of nearest/furthest neighbours $Q$, margin $\alpha$.
    \Ensure Trained model $\phi_p$, all neighbours set $\mathcal{B}$, nearest neighbours set $\mathcal{N}$, furthest neighbours set $\mathcal{F}$.
    
    \State $\mathcal{T} \gets \varnothing$, $\mathcal{B} \gets \varnothing$, $\mathcal{N} \gets \varnothing$, $\mathcal{F} \gets \varnothing$

    \For{$i \gets 1$ \textbf{to} $|\mathcal{D}|$}
      \State $a_i \gets w_i$ \hfill $\triangleright$ \text{anchor}
      \State $p_i \gets w_{i-r}$, where $r \sim \mathcal{U}(1, y)$ \hfill $\triangleright$ \text{positive pair}
      \State $n_i \gets$ \texttt{InjectAnomaly}($w_i$) \hfill $\triangleright$ \text{negative pair (Alg. \ref{algo:carlainject})}
      \State append triplet $(a_i, p_i, n_i)$ to $\mathcal{T}$ \hfill $\triangleright$ \text{triplets batches}
      \State add $a_i$ and $n_i$ to $\mathcal{B}$ \hfill $\triangleright$ \text{neighbours set with size $2|\mathcal{D}|$}
    \EndFor

    \While{\textnormal{Pretext loss} $\mathcal{L}_{pretext}(\phi_p,\mathcal{T}, \alpha)$ \textnormal{decreases}}
      \State Update $\phi_p$ with $\mathcal{L}_{pretext}$ \hfill $\triangleright$ \text{i.e. Equation (\ref{eq:pretext_loss})}
    \EndWhile

    \For{$j \gets 1$ \textbf{to} $|\mathcal{B}|$}
      \State $\mathcal{N}_j \gets Q$ nearest neighbours of $w_j \in \mathcal{B}$ in $\phi_p(\mathcal{B})$ space
      \State $\mathcal{F}_j \gets Q$ furthest neighbours of $w_j \in \mathcal{B}$ in $\phi_p(\mathcal{B})$ space
    \EndFor

    \State \textbf{Return} $\phi_p$, $\mathcal{B}$, $\mathcal{N}$, $\mathcal{F}$ \hfill $\triangleright$ \text{inputs of the next stage}
  \end{algorithmic}
\end{algorithm}

\subsubsection{Part One: Contrastive Representation Learning} 
\label{subsec3.1.1}
In this part, we introduce a contrastive learning framework for learning a discriminative representation of features for time series windows. To extract features from the time series data, we utilise a multi-channel ResNet architecture, where each channel represents a different time series dimension. Using different kernel sizes in ResNet allows us to capture features at various temporal scales, which is particularly important in analysing time series data and makes our model less sensitive to window size selection. We add an MLP layer as the final layer to produce a feature vector with lower dimensions.

To encourage the model to distinguish between different time series windows, we utilise a triplet loss function. Specifically, we create triplets of samples in the form of $(a, p, n)$, where $a$ is the anchor sample (i.e. the original time series window $w_i$), $p$ is a positive sample (i.e. another random window $w_{i-r}$, selected from $y$ previous windows, where $r \sim \mathcal{U}(1, y)$), and $n$ is a negative sample (i.e. an anomaly injected version of the original window $w_i$ to make it anomalous). Assuming we have all the triplets of $(a, p, n)$ in a set $\mathcal{T}$, the pretext triplet loss function is defined as follows:
\begin{small}
\begin{equation}
\label{eq:pretext_loss}
\mathcal{L}_{Pretext}(\phi_p, \mathcal{T}, \alpha) = \frac{1}{|\mathcal{T}|}\sum_{(a,p,n) \in \mathcal{T}}\max(\lVert \phi_p(a) - \phi_p(p) \rVert_2^2 - \lVert \phi_p(a) - \phi_p(n) \rVert_2^2 + \alpha, 0)
\end{equation}
\end{small}
Where $\phi_p(\cdot)$ is the learned feature representation neural network, $\lVert \cdot \rVert_2^2$ denotes the squared Euclidean distance, and $\alpha$ is a margin that controls the minimum distance between positive and negative samples. The objective of the pretext triplet loss function is to decrease the distance between the anchor sample and its corresponding positive sample while simultaneously increasing the distance between the anchor sample and negative samples. This approach encourages the model to learn a representation that can differentiate between normal and anomalous time series windows.

Our approach empowers the model to learn similar feature representations for temporally proximate windows. Since the majority of data is normal, the model captures temporal relationships of normal data through learning similar representations for normal windows that are temporally proximate. Furthermore, by introducing anomalies into the system, the model can learn a more effective decision boundary, resulting in a reduced false positive rate and enhanced precision in anomaly detection compared to current state-of-the-art models.

The output of Part One includes a trained neural network (ResNet) $\phi_p$ and a list of anchor and negative samples stored in $\mathcal{B}$. 

\subsubsection{Part Two: Nearest and Furthest Neighbours}
Part Two of our approach uses the feature representations learned in Part One to identify semantically meaningful nearest and furthest neighbours of each sample. To achieve this, we utilise $\mathcal{B}$ along with the indices of $Q$ nearest and $Q$ furthest neighbours of all samples in $\mathcal{B}$.

The primary goal of this part is to generate a prior that captures the \textit{semantic similarity} and \textit{semantic dissimilarity} between windows' representations using their neighbours, as our empirical analysis shows that in the majority of cases, these nearest neighbours belong to the same class (see Figure~\ref{fig:tsne-pretext}). Semantic similarity for a given window $w_i$ defines as $Q$ nearest neighbours of $\phi_p(w_i) \in \phi_p(\mathcal{B})$, where $Q$ is the number of nearest neighbours. And, semantic dissimilarity for a given window $w_i$ defined as $Q$ furthest neighbours of $\phi_p(w_i) \in \phi_p(\mathcal{B})$. 
The output of Part Two is a set of all anchor and negative samples (anomaly injected) and the indices of their $Q$ nearest neighbours $\mathcal{N}$ and $Q$ furthest neighbours $\mathcal{F}$. Utilising $\mathcal{N}$ and $\mathcal{F}$ can enhance the performance of our classification method used in the Self-supervised Classification Stage. The culmination of the Pretext stage is a comprehensive set of time series windows alongside their nearest and furthest neighbours, setting the stage for their utilisation in the forthcoming Self-supervised Classification Stage.

\subsection{Self-supervised Classification Stage}
\label{subsec3.2}
As we transition into the Self-supervised Classification Stage, we utilise the output of the Pretext stage as our foundational input. This stage includes initialising a new ResNet architecture with the learned feature representations from Part One of the Pretext Stage and then integrating the semantically meaningful nearest and furthest neighbours from Part Two as a prior ($\mathcal{N}$ and $\mathcal{F}$) into a learnable approach. Algorithm \ref{algo:carlaself} shows the steps of the Self-supervised classification Stage.

\begin{algorithm}[t]
\scriptsize
\caption{\texttt{SelfSupevisedCARLA}($\phi_p$, $\mathcal{B}$, $\mathcal{N}$, $\mathcal{F}$, $C$, $\beta$)}
\label{algo:carlaself}
\begin{algorithmic}[1]
\renewcommand{\algorithmicrequire}{\textbf{Input:}}
\renewcommand{\algorithmicensure}{\textbf{Output:}}

\Require Initial trained neural network (ResNet) from the Pretext Stage $\phi_p$, Dataset of time series windows including original windows and anomaly injected windows $\mathcal{B}$, Set of $Q$ nearest neighbours for each window $\mathcal{N}$, Set of $Q$ furthest neighbours for each window $\mathcal{F}$, Number of Classes $C$, Entropy loss weight $\beta$
\Ensure Trained model $\phi_s$, Majority class $C_m$

\State $\phi_s \gets$ initialise by $\phi_p$ \hfill $\triangleright$ \text{$\phi_p$ from Algorithm \ref{algo:carlapretext} (\texttt{PretextCARLA})}

\While{$\mathcal{L}_{Self-supervised}(\phi_s,\mathcal{B}, \mathcal{N}, \mathcal{F}, \mathcal{C}, \beta)$ \textnormal{decreases}}
  \State Update $\phi_s$ with $\mathcal{L}_{Self-supervised}$ \hfill $\triangleright$ \text{i.e. Equation (\ref{eq:selfloss})}
\EndWhile

\For{$i \gets 1$ \textbf{to} $|\mathcal{D}|$}
  \State $C^i = \arg\max(\phi_s(w_i)), w_i\in \mathcal{D}$ \hfill $\triangleright$ \text{assign class label $C_j \in \mathcal{C}$ to window $i$}
\EndFor

\State $C_m = \arg \max_{C_j \in \mathcal{C}} (n(C_j))$ \hfill $\triangleright$ \text{find majority class $C_m$}

\hfill $\triangleright$ \text{$n(C_j)$ denotes the number of members in a class $C_j$}

\State \textbf{Return} $\phi_s$, $C_m$
\end{algorithmic}
\end{algorithm}

To encourage the model to produce both consistent and discriminative predictions, we employ a contrastive approach with a customised loss function. Specifically, the loss function maximises the similarity between each window representation and its nearest neighbours while minimising the similarity between each window representation and its furthest neighbours. The loss function can be defined as follows: At the beginning of this stage, we have $\mathcal{B}$ from the Pretext Stage, which is the set of all original window representations (we call them anchors in this stage) and their corresponding anomalous representations. Let $C$ be the number of classes. We also have the $Q$ nearest neighbours of the anchor samples $\mathcal{N}$ and the $Q$ furthest neighbours of the anchor samples $\mathcal{F}$. 

We aim to learn a classification neural network function $\phi_s$ - initialised by $\phi_p$ from the Pretext Stage - that classifies a window $w_i$ and its $Q$ nearest neighbours to the same class, and $w_i$ and its $Q$ furthest neighbours to different classes. The neural network $\phi_s$ terminates in a softmax function to perform a soft assignment over the classes $\mathcal{C}=\{1, ..., C\}$, with $\phi_s(w) \in [0,1]^C$.

To encourage similarity between the anchor samples and their nearest neighbours, we compute the pairwise similarity between the probability distributions of the anchor and its neighbours as it is shown in Equation (\ref{eq:similarity}). The dot product between an anchor and its neighbour will be maximal when the output of the softmax for them is close to 1 or 0 and consistent in that it is assigned to the same class. Then, we define a consistency loss in Equation (\ref{eq:consistencyloss}) using the binary cross entropy to maximise the similarity between the anchor and nearest neighbours.
\begin{small}
    \begin{equation}
        \label{eq:similarity}
        \text{similarity}(\phi_s, w_i, w_j) = \langle \phi_s(w_i) \cdot \phi_s(w_j) \rangle
        = \phi_s(w_i)^\top \phi_s(w_j) 
\end{equation}
\end{small}
\vspace{-15pt}
\begin{small}
\begin{equation}
\label{eq:consistencyloss}
\mathcal{L}_{consistency}(\phi_s,\mathcal{B}, \mathcal{N}) =
-\frac{1}{\lvert \mathcal{B} \rvert} \sum_{w \in \mathcal{B}} \sum_{w_n \in \mathcal{N}_{w}} \log(\text{similarity}(\phi_s, w, w_n))
\end{equation}
\end{small}
The consistency loss aims to strengthen the alignment of anchor samples with their nearest neighbours, promoting cohesion by enhancing the similarity within neighbours.

We define an inconsistency loss to encourage dissimilarity between the anchor samples and their furthest neighbours in Equation (\ref{eq:inconsistencyloss}). In this regard, we compute the pairwise similarity between the probability distributions of the anchor and furthest neighbour samples as well. Then, use the binary cross entropy loss to minimise the similarity to the furthest neighbours in the final loss function. While the mathematical form mirrors the consistency loss, its application diverges. Here, the similarity measure is used inversely; we seek to minimise this similarity, thereby driving a distinction between the anchor and its furthest neighbours to underscore cluster separation.

\begin{small}
\begin{equation}
\label{eq:inconsistencyloss}
\mathcal{L}_{inconsistency}(\phi_s,\mathcal{B}, \mathcal{F}) =
-\frac{1}{\lvert \mathcal{B} \rvert} \sum_{w \in \mathcal{B}} \sum_{w_n \in \mathcal{F}_{w}} \log(\text{similarity}(\phi_s, w, w_n))
\end{equation}
\end{small}

To encourage class diversity and prevent overfitting, we apply entropy loss on the distribution of anchor and neighbour samples across classes. Assuming the classes set is denoted as $\mathcal{C}=\{1,...,C\}$, and the probability of window $w_i$ being assigned to class $c$ is denoted as $\phi_s^c(w_i)$:
\begin{small}
\begin{equation}
\mathcal{L}_{entropy}(\phi_s,\mathcal{B}, \mathcal{C}) = \sum_{c \in \mathcal{C}} \hat{\phi_s^c} \log(\hat{\phi_s^c}) \\
\text{ where } \hat{\phi_s^c}
= \frac{1}{\lvert \mathcal{B} \rvert}\sum_{w_i \in \mathcal{B}} \phi_s^c(w_i)
\end{equation}
\end{small}

The final objective in the Self-supervised Classification Stage is to minimise the total loss, calculated by the difference between the consistency and inconsistency losses, reduced by the entropy loss multiplied by a weight parameter, $\beta$:
\begin{small}
\begin{equation}
\label{eq:selfloss}
\begin{aligned}
& \mathcal{L}_{Self-supervised}(\phi_s,\mathcal{B}, \mathcal{N}, \mathcal{F}, \mathcal{C}, \beta) = \\
& \mathcal{L}_{consistency}(\phi_s,\mathcal{B}, \mathcal{N}) - \mathcal{L}_{inconsistency}(\phi_s,\mathcal{B}, \mathcal{F}) - \beta \cdot \mathcal{L}_{entropy}(\phi_s,\mathcal{B}, \mathcal{C})
\end{aligned}
\end{equation}
\end{small}

The goal of the loss function is to learn a representation that is highly discriminative, with the nearest neighbours assigned to the same class as the anchor samples and the furthest neighbours assigned to a distinct class. By incorporating the semantically meaningful nearest and furthest neighbours, the model is able to produce more consistent and confident predictions, with the probability of a window being classified as one particular class is close to 1 or 0.

Overall, the loss function $\mathcal{L}_{Self-supervised}$ represents a critical component of our approach to time series anomaly detection, as it allows us to effectively learn a discriminative feature representation that can be utilised to distinguish between normal and anomalous windows.

\subsection{CARLA's Inference} 
\label{subsec3.3}
Upon completing Part Two of our approach, we determine the class assignments for set $\mathcal{D}$ and majority class $C_m$, where  $C_m = \arg \max_{C_j \in \mathcal{C}} (n(C_j))$ which comprises the class with the highest number of anchors (i.e. original windows). For every new window $w_t$ during inference, we calculate $\phi_s^{C_m}(w_t)$, representing the probability of window $w_t$ being assigned to the majority class $C_m$. Using an end-to-end approach, we classify a given window $w_t$ as normal or anomalous based on whether it belongs to the majority class. Specifically, Equation (\ref{eq:anomalylabel}) describes how we infer a label for a new time series window $w_t$. Additionally, we can employ Equation (\ref{eq:anomalyscore}) to generate an anomaly score of $w_t$ for further analysis.
\begin{small}
\begin{equation}
\label{eq:anomalylabel}
\text{Anomaly label } (w_t): \begin{cases}
0, & \text{if } \forall c \in \mathcal{C},\phi_s^{C_m}(w_t) \geq \phi_s^{c}(w_t) \\
1, & \text{otherwise}
\end{cases}
\end{equation}
\end{small}

\begin{small}
\begin{equation}
\label{eq:anomalyscore}
\text{Anomaly score } (w_t): 1-\phi_s^{C_m}(w_t)
\end{equation}
\end{small}

By assigning a window to a specific class, our model can determine the likelihood of a given window being normal or anomalous with greater precision.

Furthermore, the probability $\phi_s^{C_m}(w_t)$ can be used to calculate an anomaly score, with lower values indicating a higher probability of anomalous behaviour. This score can be useful for further analysis and can aid in identifying specific characteristics of anomalous behaviour.

\section{Experiments} \label{sec4}
The objective of this section is to thoroughly assess CARLA's performance through experiments conducted on multiple benchmark datasets and to compare its results with multiple alternative methods. Section \ref{subsec4.1} provides an overview of the benchmark datasets used in the evaluation, highlighting their significance in assessing the effectiveness of our model. Section \ref{subsec4.2} delves into the benchmark methods employed for comparing the performance of different models. In section \ref{subsec4.3}, we discuss the evaluation setup, including the hyper-parameters chosen for our approach. Moving forward, section \ref{subsec4.4} provides results for all benchmark methods based on the respective benchmark datasets, facilitating a comprehensive comparison. Additionally, we explore the behaviour of CARLA across diverse data configurations and variations in the ablation studies (section \ref{subsec4.5}) and investigate its sensitivity to parameter changes in section \ref{subsec4.6}. All evaluations were conducted on a system equipped with an A40 GPU, 13 CPUs, and 250 GB of RAM.

\setlength{\tabcolsep}{3pt}
\begin{table}[t]
\centering
\scriptsize
\caption{Statistics of the benchmark datasets used.}
\label{tab:dss}
\begin{tabular}{c|cccccc}
\hline
\textbf{Benchmark} & \textbf{\# datasets} & \textbf{\# dims} & \textbf{Train size} & \textbf{Test size} & \textbf{Anomaly\%} \\ \hline
MSL         & 27 & 55 & 58,317  & 73,729  & 10.72\% \\
SMAP       & 55 & 25 & 140,825 & 444,035 & 13.13\% \\
SMD         & 28 & 38 & 708,405 & 708,420 & 4.16\%  \\
SWaT         & 1 & 51 & 496,800 & 449,919 & 12.33\%  \\
WADI         & 1 & 123 & 784,568 & 172,801 & 5.77\%  \\
Yahoo-A1 & 67 & 1  & 46,667  & 46,717  & 1.76\%  \\
KPI & 29 & 1  & 1,048,576 & 2,918,847 & 1.87\%  \\ \hline
\end{tabular}%
\end{table}

\subsection{Benchmark Datasets} \label{subsec4.1}
We evaluate the performance of the proposed model and make comparisons of the results across the seven most commonly used real benchmark datasets for TSAD. The datasets are summarised in Table~\ref{tab:dss}, which provides key statistics for each dataset. All datasets, except Yahoo, have a predefined train/test split with unlabeled training data.\\
%\begin{itemize}
\textbf{NASA Datasets} -- \textbf{Mars Science Laboratory (MSL)} and \textbf{Soil Moisture Active Passive (SMAP)}\footnote{\url{https://www.kaggle.com/datasets/patrickfleith/nasa-anomaly-detection-dataset-smap-msl}} --~\citep{hundman2018detecting} These real-world datasets, collected from NASA spacecraft, contain anomaly information from incident reports for a spacecraft monitoring system.\\
\textbf{Server Machine Dataset (SMD)}\footnote{\url{https://github.com/NetManAIOps/OmniAnomaly/tree/master/ServerMachineDataset}}~\citep{su2019robust} is gathered from 28 servers, incorporating 38 sensors, over a span of 10 days. During this period, normal data was observed within the initial 5 days, while anomalies were sporadically injected during the subsequent 5 days.\\
\textbf{Secure water treatment (SWaT)}\footnote{\url{https://itrust.sutd.edu.sg/testbeds/secure-water-treatment-swat/}}~\citep{mathur2016swat} is meticulously gathered over 11 days, utilising a water treatment platform with 51 sensors. During the initial 7 days, it consisted exclusively of normal data, reflecting regular operational conditions. In the following 4 days, 41 anomalies were deliberately generated using a wide range of attacks.\\
\textbf{Water distribution testbed (WADI)}\footnote{\url{https://itrust.sutd.edu.sg/testbeds/water-distribution-wadi/}}~\citep{ahmed2017wadi} is acquired from a scaled-down urban water distribution system that included a combined total of 123 actuators and sensors. This data covers a span of 16 days. Notably, the final two days of the dataset specifically feature anomalies. In addition, the test dataset encompasses 15 distinct anomaly segments, providing valuable examples for anomaly detection and analysis purposes.\\
\textbf{Yahoo}\footnote{\url{https://webscope.sandbox.yahoo.com/catalog.php?datatype=s&did=70}} dataset~\citep{yahoods} is a benchmark dataset that consists of hourly 367 sampled time series with anomaly points labels. The dataset includes four benchmarks, with the A1 benchmark containing "real" production traffic data from Yahoo properties, and the other three are synthetic. For evaluation, we focus on the A1 benchmark, which includes 67 univariate time series. We use half of each dataset for training and the remaining for test, excluding those without anomalies in the test set.\\
\textbf{Key performance indicators (KPI)}\footnote{\url{https://github.com/NetManAIOps/KPI-Anomaly-Detection}} contains data of service and machine key performance indicators (KPIs) from real scenarios of Internet companies. Service KPIs measure web service performance, such as page response time and page views. Machine KPIs reflect machine health, including CPU and memory utilisation.

\subsection{Benchmark Methods} \label{subsec4.2}
We provide a description of the ten prominent TSAD models that were used for comparison with CARLA, spanning various TSAD categories. For the semi-supervised approach, the LSTM-VAE~\citep{park2018multimodal} is highlighted, which is a point-wise anomaly detection model. In the realm of unsupervised reconstruction-based models, Donut~\citep{xu2018unsupervised} stands out for UTS, while OmniAnomaly~\cite{su2019robust} and AnomalyTransformer~\citep{xu2021anomaly} are noted for MTS, alongside TranAD~\citep{Tuli2022TranADDT} which caters to both UTS and MTS. The unsupervised forecasting-based model, THOC~\cite{shen2020timeseries} and TimesNet~\cite{wu2023timesnet}, are recognised for their predictive capabilities. Within the self-supervised category, MTAD-GAT~\citep{zhao2020multivariate} emerges as a hybrid model, with TS2Vec~\cite{yue2022ts2vec} and DCdetector~\cite{yang2023dcdetector} offering representation-based methodologies. Additionally, recent advancements are represented by TimesNet~\cite{wu2023timesnet} and DCdetector~\cite{yang2023dcdetector}, showcasing the ongoing evolution in TSAD models.
The ``Random Anomaly Score'' model is designed to generate anomaly scores based on a normal distribution ($\mathcal{N}(0, 1)$) with mean 0 and standard deviation 1. This model is specifically developed to illustrate evaluation metrics and strategies in the field of TSAD.

\subsection{Evaluation Setup} \label{subsec4.3}
In our study, we evaluate ten TSAD models on benchmark datasets previously mentioned in section \ref{subsec4.1} using their best hyper-parameters, as they stated, to ensure a fair evaluation. We summarise the default hyper-parameters used in our implementation as follows: The CARLA model consists of a 3-layer ResNet architecture with three different kernel sizes [8, 5, 3] to capture temporal dependencies. The dimension of the representation is 128. We use the same hyper-parameters across all datasets to evaluate CARLA: window size = 200, number of classes = 10, number of nearest/furthest neighbours ($Q$) = 5, and coefficient of entropy = 5. For detailed information about all experiments involving the aforementioned hyper-parameter choices, please refer to the section \ref{subsec4.5}. We run the Pretext Stage for 30 epochs and the Self-supervised Classification Stage for 100 epochs on all datasets. 

It is important to note that we do not use Point Adjustment (PA) in our evaluation process. Despite its popularity,~\cite{kim2022towards} found that applying PA leads to an overestimation of TSAD models' capability and can bias results towards methods that produce extreme anomaly scores. To ensure accuracy, we present conventional F1 scores and relegate PA results to \ref{PA-section}. The F1 score without PA is referred to as F1. %, while adjusted predictions are denoted as $F1_{PA}$ in  \footnote{\url{??}}.

\begin{table*}[t!]
\centering
\scriptsize
\caption{Precision (Prec), recall (Rec), F1 and AU-PR results for various models on multivariate time series datasets. The best results are in bold, and the second-best is indicated by \underline{underline}. Due to the single time series in both the SWaT and WADI datasets, the standard deviation of AU-PR is not available.}
\label{tab:mts_compare}
\begin{tabular}{c|c|ccccc||c}
\hline
Model & Metric & MSL & SMAP & SMD & SWaT & WADI & Avg Rank \\
\hline
\multirow{4}{*}{\begin{tabular}[c]{@{}c@{}}LSTM-VAE~\cite{park2018multimodal} \\ (2018)\end{tabular}} & Prec & 0.2723 & 0.2965 & 0.2045 & 0.9707 & 0.0596 & \multirow{4}{*}{5.5}\\
& Rec & 0.8083 & 0.8303 & 0.5491 & 0.5769 & 1.0000 \\
& F1 & 0.4074 & 0.4370 & 0.2980 & 0.7237 & 0.1126 \\
& AU-PR & 0.285$\pm$0.249 & 0.258$\pm$0.305 & 0.395$\pm$0.257 & \underline{0.685} & 0.039 \\
\hline
\multirow{4}{*}{\begin{tabular}[c]{@{}c@{}}OmniAnomaly~\cite{su2019robust} \\ (2019) \end{tabular}} & Prec & 0.1404 & 0.1967 & 0.3067 & 0.9068 & 0.1315 & \multirow{4}{*}{5.0}\\
& Rec & 0.9085 & 0.9424 & 0.9126 & 0.6582 & 0.8675 \\
& F1 & 0.2432 & 0.3255 & \underline{0.4591} & \textbf{0.7628} & 0.2284 \\
& AU-PR & 0.149$\pm$0.182 & 0.115$\pm$0.129 & 0.365$\pm$0.202 & \textbf{0.713} & 0.120 \\
\hline
\multirow{4}{*}{\begin{tabular}[c]{@{}c@{}} MTAD-GAT~\cite{zhao2020multivariate} \\ (2020) \end{tabular}} & Prec & 0.3559 & 0.3783 & 0.2473 & 0.1387 & 0.0706 & \multirow{4}{*}{4.2}\\
& Rec & 0.7067 & 0.8239 & 0.5834 & 0.9585 & 0.5838 \\
& F1 & \underline{0.4734} & \underline{0.5186} & 0.3473 & 0.2423 & 0.1259 \\
& AU-PR & \underline{0.335$\pm$0.259} & \underline{0.339$\pm$0.300} & 0.401$\pm$0.263 & 0.095 & 0.084 \\
\hline
\multirow{4}{*}{\begin{tabular}[c]{@{}c@{}} THOC~\cite{shen2020timeseries} \\ (2020) \end{tabular}} & Prec & 0.1936 & 0.2039 & 0.0997 & 0.5453 & 0.1017 & \multirow{4}{*}{6.8}\\
& Rec & 0.7718 & 0.8294 & 0.5307 & 0.7688 & 0.3507 \\
& F1 & 0.3095 & 0.3273 & 0.1679 & 0.6380 & 0.1577 \\
& AU-PR & 0.239$\pm$0.273 & 0.195$\pm$0.262 & 0.107$\pm$0.126 & 0.537 & 0.103 \\
\hline
\multirow{4}{*}{\begin{tabular}[c]{@{}c@{}} AnomalyTran~\cite{xu2021anomaly} \\ (2021) \end{tabular} } & Prec & 0.2182 & 0.2669 & 0.2060 & 0.9711 & 0.0601 & \multirow{4}{*}{5.4}\\
& Rec & 0.8231 & 0.8600 & 0.5822 & 0.5946 & 0.9604 \\
& F1 & 0.3449 & 0.4074 & 0.3043 & \underline{0.7376} & 0.1130 \\
& AU-PR & 0.236$\pm$0.237 & 0.264$\pm$0.315 & 0.273$\pm$0.232 & 0.681 & 0.040 \\
\hline
\multirow{4}{*}{\begin{tabular}[c]{@{}c@{}}TranAD~\cite{Tuli2022TranADDT} \\ (2022) \end{tabular}} & Prec & 0.2957 & 0.3365 & 0.2649 & 0.1927 & 0.0597 & \multirow{4}{*}{6.2}\\
& Rec & 0.7763 & 0.7881 & 0.5661 & 0.7965 & 1.0000 \\
& F1 & 0.4283 & 0.4716 & 0.3609 & 0.3103 & 0.1126 \\
& AU-PR & 0.278$\pm$0.239 & 0.287$\pm$0.300 & \underline{0.412$\pm$0.260} & 0.192 & 0.039 \\
\hline
\multirow{4}{*}{\begin{tabular}[c]{@{}c@{}} TS2Vec~\cite{yue2022ts2vec} \\ (2022) \end{tabular}} & Prec & 0.1832 & 0.2350 & 0.1033 & 0.1535 & 0.0653 & \multirow{4}{*}{7.7}\\
& Rec & 0.8176 & 0.8826 & 0.5295 & 0.8742 & 0.7126 \\
& F1 & 0.2993 & 0.3712 & 0.1728 & 0.2611 & 0.1196 \\
& AU-PR & 0.132$\pm$0.135 & 0.148$\pm$0.165 & 0.113$\pm$0.075 & 0.136 & 0.057 \\
\hline
\multirow{4}{*}{\begin{tabular}[c]{@{}c@{}} DCdetector~\cite{yang2023dcdetector} \\ (2023) \end{tabular}} & Prec & 0.1288 & 0.1606 & 0.0432 & 0.1214 & 0.1417 & \multirow{4}{*}{9.1}\\
& Rec & 0.9578 & 0.9619 & 0.9967 & 0.9999 & 0.9684 \\
& F1 & 0.2270 & 0.2753 & 0.0828 & 0.2166 & \underline{0.2472} \\
& AU-PR & 0.129$\pm$0.144 & 0.124$\pm$0.153 & 0.043$\pm$0.036 & 0.126 & \underline{0.121} \\
\hline
\multirow{4}{*}{\begin{tabular}[c]{@{}c@{}} TimesNet~\cite{wu2023timesnet} \\ (2023) \end{tabular}} & Prec & 0.2257 & 0.2587 & 0.2450 & 0.1214 & 0.1334 & \multirow{4}{*}{6.1}\\
& Rec & 0.8623 & 0.8994 & 0.5474 & 1.0000 & 0.1565 \\
& F1 & 0.3578 & 0.4019 & 0.3385 & 0.2166 & 0.1440 \\
& AU-PR & 0.283$\pm$0.213 & 0.208$\pm$0.211 & 0.385$\pm$0.225 & 0.083 & 0.084 \\
\hline
\multirow{4}{*}{Random} & Prec & 0.1746 & 0.1801 & 0.0952 & 0.1290 & 0.0662 & \multirow{4}{*}{8.4}\\
& Rec & 0.9220 & 0.9508 & 0.9591 & 0.9997 & 0.9287 \\
& F1 & 0.2936 & 0.3028 & 0.1731 & 0.2166 & 0.1237 \\
& AU-PR & 0.172$\pm$0.133 & 0.140$\pm$0.148 & 0.089$\pm$0.058 & 0.129 & 0.067 \\
\hline
\multirow{4}{*}{CARLA} & Prec & 0.3891 & 0.3944 & 0.4276 & 0.9886 & 0.1850 & \multirow{4}{*}{1.6}\\
& Rec & 0.7959 & 0.8040 & 0.6362 & 0.5673 & 0.7316 \\
& F1 & \textbf{0.5227} & \textbf{0.5292} & \textbf{0.5114} & 0.7209 & \textbf{0.2953} \\
& AU-PR & \textbf{0.501$\pm$0.267} & \textbf{0.448$\pm$0.326} & \textbf{0.507$\pm$0.195} & 0.681 & \textbf{0.126} \\
\hline
\end{tabular}
\end{table*}

\begin{table}[t]
    \centering
    \scriptsize
    \caption{Precision (Prec), recall (Rec), F1, and AU-PR results for various models on univariate time series datasets. The best results are in bold, and the second-best is indicated by \underline{underline}.}
    \label{tab:uts_compare}
    \begin{tabular}{c|c|cc||c}
        \hline
        Model & Metric & Yahoo-A1 & KPI & Avg Rank \\
        \hline
        \multirow{4}{*}{\begin{tabular}[c]{@{}c@{}} Donut~\cite{xu2018unsupervised} \\ (2018) \end{tabular}} & Prec & 0.3239 & 0.0675 & \multirow{4}{*}{5.0}  \\
                           & Rec & 0.9955 & 0.9340 &  \\
                           & F1 & 0.4888 & 0.1259 &  \\
                           & AU-PR & 0.264$\pm$0.244 & 0.078$\pm$0.073 & \\
        \hline
        \multirow{4}{*}{\begin{tabular}[c]{@{}c@{}} THOC~\cite{shen2020timeseries} \\ (2020) \end{tabular}} & Prec & 0.1495 & 0.1511 & \multirow{4}{*}{5.5}  \\
                          & Rec & 0.8326 & 0.5116 &  \\
                          & F1 & 0.2534 & 0.2334 &  \\
                          & AU-PR & 0.349$\pm$0.342 & 0.229$\pm$0.217 & \\
        \hline
        \multirow{4}{*}{\begin{tabular}[c]{@{}c@{}} TranAD~\cite{Tuli2022TranADDT} \\ (2022) \end{tabular}} & Prec & 0.4185 & 0.2235 & \multirow{4}{*}{2.0}  \\
                        & Rec & 0.8712 & 0.4016 &  \\
                        & F1 & \underline{0.5654} & \underline{0.2872} &  \\
                        & AU-PR & \textbf{0.691$\pm$0.324} & \underline{0.285$\pm$0.206} &  \\
        \hline
        \multirow{4}{*}{\begin{tabular}[c]{@{}c@{}} TS2Vec~\cite{yue2022ts2vec} \\ (2022) \end{tabular}} & Prec & 0.3929 & 0.1333 & \multirow{4}{*}{5.0}  \\
                        & Rec & 0.6305 & 0.4329 &  \\
                        & F1 & 0.4841 & 0.2038 &  \\
                        & AU-PR & 0.491$\pm$0.352 & 0.221$\pm$0.156 &  \\
        \hline
        \multirow{4}{*}{\begin{tabular}[c]{@{}c@{}} DCdetector~\cite{yang2023dcdetector} \\ (2023) \end{tabular}} & Prec & 0.0598 & 0.0218 & \multirow{4}{*}{8.0}  \\
                        & Rec & 0.9434 & 0.8589 &  \\
                        & F1 & 0.1124 & 0.0425 &  \\
                        & AU-PR & 0.041$\pm$0.059 & 0.018$\pm$0.017 &  \\
        \hline
        \multirow{4}{*}{\begin{tabular}[c]{@{}c@{}} TimesNet~\cite{wu2023timesnet} \\ (2023) \end{tabular}} & Prec & 0.3808 & 0.2174 & \multirow{4}{*}{3.0}  \\
                        & Rec & 0.7883 & 0.2713 &  \\
                        & F1 & 0.5135 & 0.2414 &  \\
                        & AU-PR & \underline{0.671$\pm$0.307} & 0.237$\pm$0.148 &  \\
        \hline
        \multirow{4}{*}{Random} & Prec & 0.2991 & 0.0657 & \multirow{4}{*}{6.5}\\
                        & Rec & 0.9636 & 0.9488 &  \\
                        & F1 & 0.4565 & 0.1229 &  \\
                        & AU-PR & 0.258$\pm$0.184 & 0.079$\pm$0.064 & \\
        \hline
        \multirow{4}{*}{CARLA} & Prec & 0.5747 & 0.1950 & \multirow{4}{*}{1.0} \\
                       & Rec & 0.9755 & 0.7360 &  \\
                       & F1 & \textbf{0.7233} & \textbf{0.3083} &  \\
                       & AU-PR & 0.645$\pm$0.352 & \textbf{0.299$\pm$0.245} &  \\
        \hline
    \end{tabular}
\end{table}

\subsection{Benchmark Comparison} \label{subsec4.4}
The performance metrics employed to evaluate the effectiveness of all models consist of precision, recall, the traditional F1 score (F1 score without PA), FPR and the area under the precision-recall curve (AU-PR). Additionally, we computed the average ranks of the models based on their F1. For all methods, we used the precision-recall curve on the anomaly score for each time series in the datasets to find the best F1 score based on precision and recall for the target time series. 

Since certain benchmark datasets such as MSL contain multiple time series datasets (as shown in Table \ref{tab:dss}), we cannot merge or combine these time series due to the absence of timestamp information. Additionally, calculating the F1 score for the entire dataset by averaging individual scores is not appropriate. In terms of precision and recall, the F1 score represents the harmonic mean, which makes it a non-additive metric. To address this, we get the confusion matrix for each time series, i.e., we calculate the number of true positives (TP), false positives (FP), true negatives (TN), and false negatives (FN) for each dataset of a benchmark such as MSL. Then, we sum up the TP, FP, TN, and FN values from all the confusion matrices to get an overall confusion matrix of the entire benchmark dataset. After that, we use the overall confusion matrix to calculate the overall precision, recall, and F1 score. This way, we ensure that the F1 score is calculated correctly for the entire dataset rather than being skewed by averaging individual F1 scores.

Furthermore, for datasets with more than one time series, we report the average and standard deviation of AU-PR across all time series. AU-PR is advantageous in imbalanced datasets because it is less sensitive to the distribution of classes~\cite{saito2015precision}. It considers the trade-off between precision and recall across all possible decision thresholds, making it more robust in scenarios where the number of instances in classes is imbalanced. By focusing on the positive class (i.e. anomalous) and its predictions, AU-PR offers a more precise evaluation of a model's ability to identify and prioritise the minority class correctly.
\begin{figure}[t]
    \centering
    \begin{subfigure}[b]{0.49\columnwidth}
        \centering
        \includegraphics[width=\textwidth]{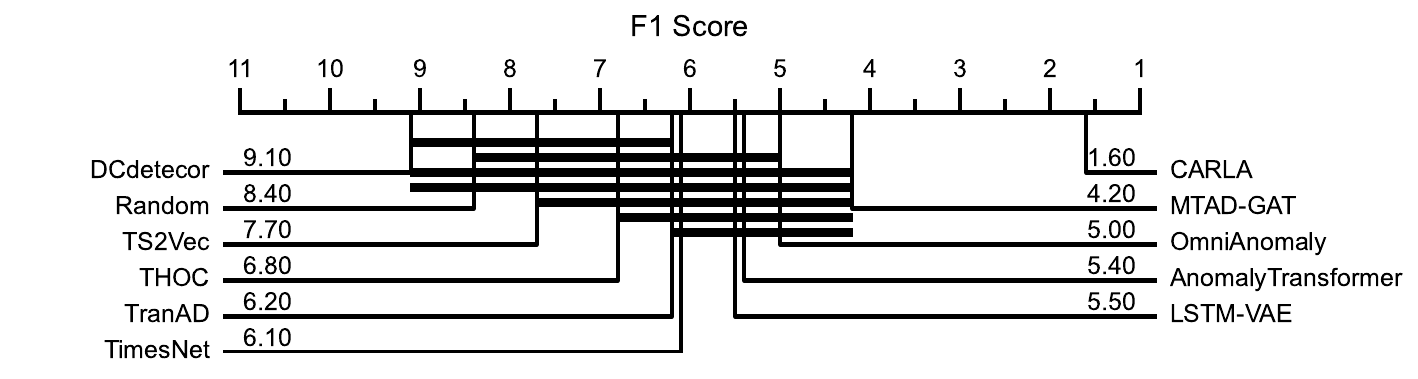}
        \caption{}
        \label{fig:cdd-m}
    \end{subfigure}
    \begin{subfigure}[b]{0.49\columnwidth}
        \centering
        \includegraphics[width=\textwidth]{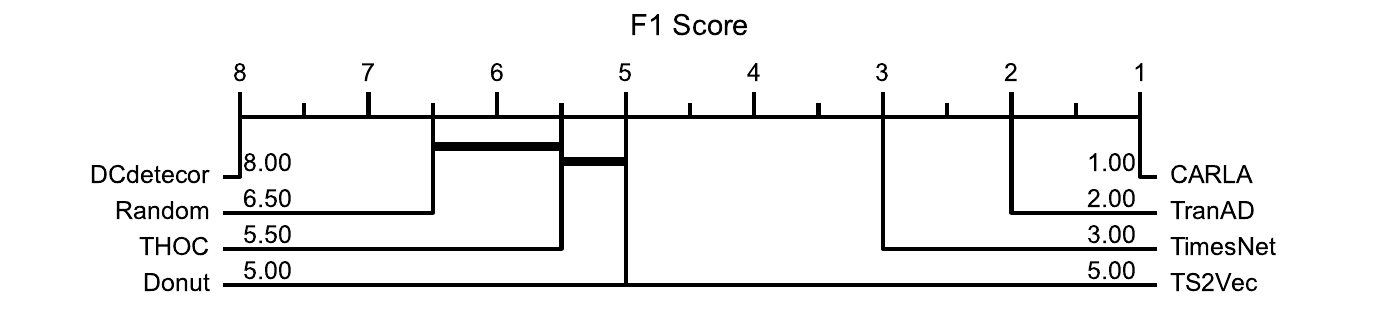}
        \caption{}
        \label{fig:cdd-u}
    \end{subfigure}
    \caption{Critical difference diagrams for multivariate (\subref{fig:cdd-m}) and univariate (\subref{fig:cdd-u}) time series benchmark datasets. }
    \label{fig:CDD}
\end{figure}
Tables \ref{tab:mts_compare} and \ref{tab:uts_compare} demonstrate the performance comparison between CARLA and all benchmark methods for multivariate and univariate datasets, respectively. These two tables show our model outperforms other models with higher F1 and AU-PR scores across all datasets (except SWaT for F1 and AU-PR and Yahoo-A1 for AU-PR), in which CARLA is the third best. This shows the strength of CARLA in generalising normal patterns due to its high precision. CARLA's precision is the highest across all seven datasets. At the same time, CARLA has high enough recall, and as a result, our CARLA's F1 is the highest across all datasets except SWaT. While the recall is higher in other benchmarks, such as OmniAnomaly, it is with the expense of very low precision (with median precision $<19.67\%$ across all datasets). This means that a good balance between precision and recall is achieved by CARLA. This is also shown in our consistently high AU-PR compared to other models and indicates that it is proficient in accurately identifying anomalous instances with higher recall while minimising false alarms with higher precision on imbalanced datasets where normal instances are predominant. Finally, CARLA achieved the lowest average rank (best rank), based on F1, in both multivariate and univariate benchmarks (see last column in Tables~\ref{tab:mts_compare} and \ref{tab:uts_compare}).

Figure~\ref{fig:CDD} shows the average ranking for each model in a critical difference diagram~\cite{demvsar2006statistical}. Models within the same clique (black bar) are not statistically significant. Statistically, the Wilcoxon signed-rank test with Holm correction is used as a post hoc test to Friedman's test. The diagrams cover 208 univariate and multivariate datasets (detailed in Table~\ref{tab:dss}). CARLA consistently ranks as the best-performing model, scoring 1.00 in univariate datasets (Figure~\ref{fig:cdd-u}) and 1.60 in multivariate datasets (Figure~\ref{fig:cdd-m}), highlighting its superiority. Moreover, CARLA is significantly different from other models.
\begin{figure*}[t!]
    \centering
    \includegraphics[width=0.9\linewidth]{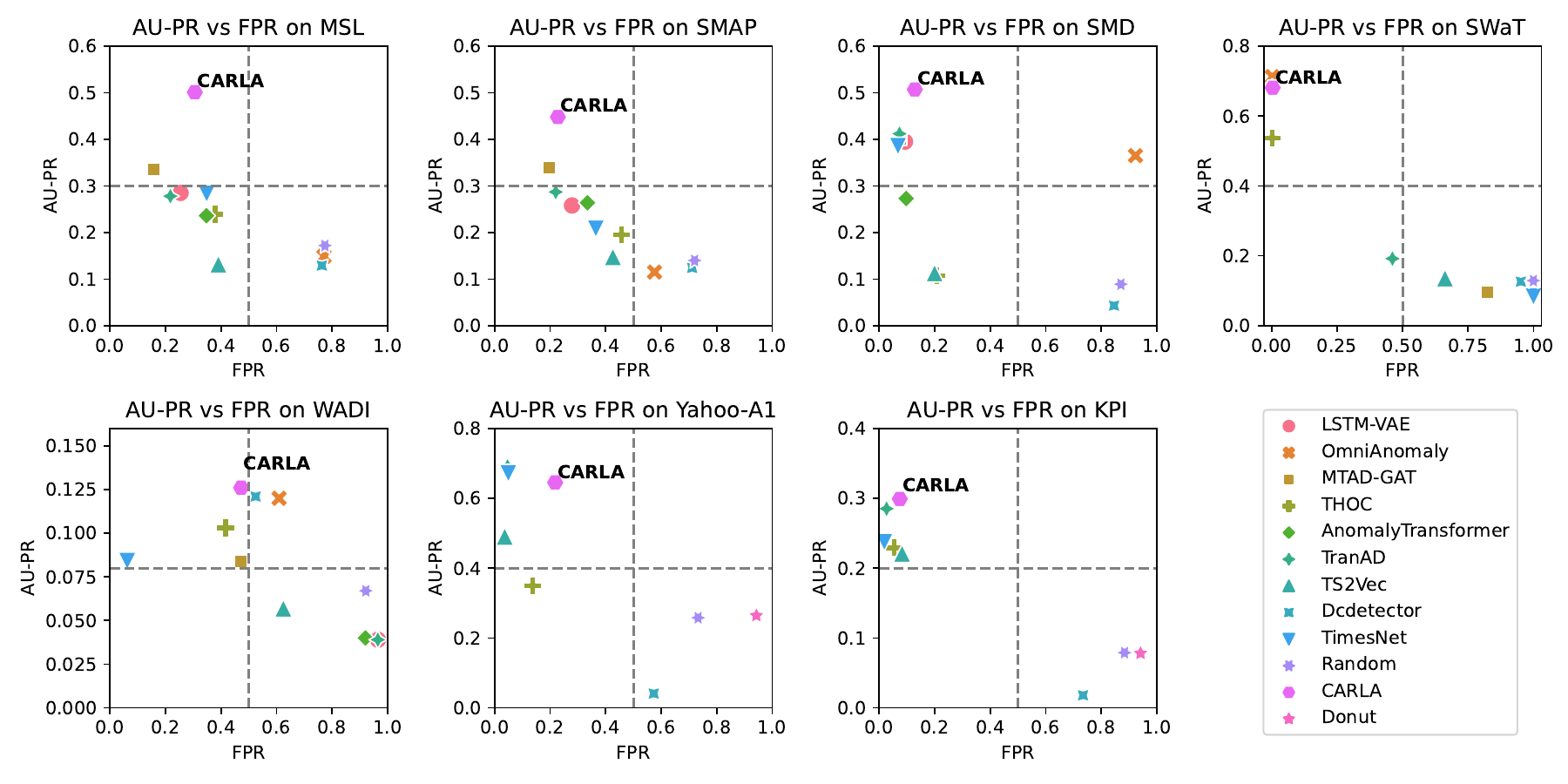}
    \caption {Models' performance comparison on MTS \& UTS datasets regarding their FPR and AU-PR. Models in the optimal quadrant (low FPR, high AU-PR) have superior anomaly detection with minimal false alarms.}
    \label{fig:fpr-aupr-multi}
\end{figure*}

We now evaluate the performance of all models in terms of FPR. We also incorporate AU-PR in this comparison, as some models can achieve a low FPR at the expense of lowering AU-PR. In the scatter plots provided in Figure~\ref{fig:fpr-aupr-multi}, we show FPR on the x-axis and AU-PR on the y-axis. The ideal point on this scatter plot is in the leftmost top corner. CARLA consistently shows a lower FPR compared to other models across various multivariate time series datasets. This is particularly noteworthy in the context of anomaly detection, where the cost of false positives can be substantial, leading to wasted resources and potential overlook of true anomalies. CARLA's position in the plots is close to the leftmost top corner (in the optimal quadrant). This indicates fewer false alarms while maintaining competitive AU-PR scores, which measure the precision and recall balance of anomaly detection.

For instance, on the MSL dataset, CARLA is among the models with the lowest FPR, and its AU-PR score is in the upper tier, showcasing its ability to identify true anomalies accurately. On the SMAP and SMD datasets, CARLA maintains a significantly lower FPR than most models, suggesting its robustness to varying data conditions. In the SWaT dataset, CARLA excels with the lowest FPR and the third highest AU-PR, marking it suitable for precision-critical applications. Even on WADI, while CARLA's FPR is slightly higher, it remains in the optimal quadrant, reflecting a strong FPR and AU-PR balance vital for operational continuity and effective anomaly detection.

Adding to CARLA's impressive performance on multivariate datasets, its capabilities in univariate time series analysis, as shown in Figure~\ref{fig:fpr-aupr-multi}, further highlighting its robustness. On the Yahoo-A1 dataset, while CARLA does not achieve the highest AU-PR, it still demonstrates a commendable balance of a low FPR and a high AU-PR, positioning it as the third-best model in this regard. Its place in the upper left quadrant indicates a strong ability to correctly identify anomalies with a minimal rate of false alarms. On the KPI dataset, CARLA maintains a competitive edge with a low FPR that surpasses most other models. While the AU-PR is the highest, CARLA's ability to limit false positives is a notable strength.

Overall, the consistent performance of CARLA across multivariate and univariate datasets and its proficiency in dealing with a variety of data types, as shown by its low FPR, highlights its strength as a reliable model for anomaly detection in MTS and UTS data. Its ability to minimise false positives without significantly sacrificing true positive detection makes it an excellent choice for scenarios where high-confidence alerts are vital.

\subsection{Ablation Study} \label{subsec4.5}
For a deeper understanding of the contributions of different stages and components in our proposed model, we conduct an ablation study. Our analysis focused on: (i) \textbf{Effectiveness of CARLA's two stages}, (ii) \textbf{Positive pair selection strategy} (iii) \textbf{Effectiveness of Different Anomaly Types} (iv) \textbf{Effectiveness of the loss components}.

\subsubsection{Effectiveness of CARLA's Stages}
\begin{figure*}[t]
\centering
    \begin{subfigure}[b]{0.23\textwidth}
        \centering
        \includegraphics[width=\textwidth]{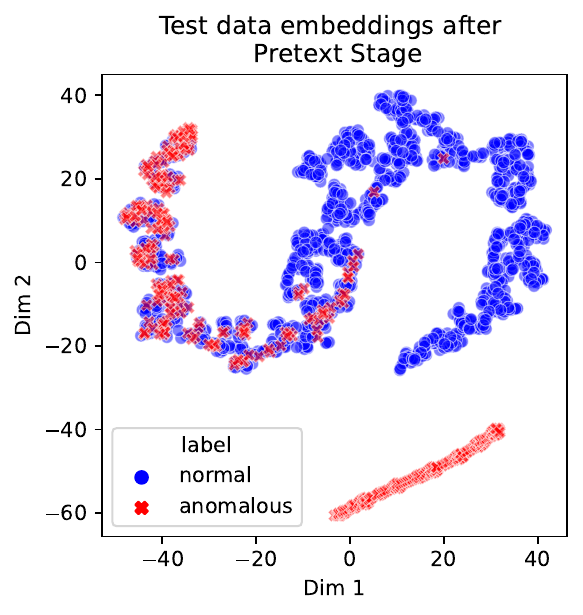}
        \caption{}
        \label{fig:tsne-pretext}
    \end{subfigure}
    \begin{subfigure}[b]{0.23\textwidth}
        \centering
        \includegraphics[width=\textwidth]{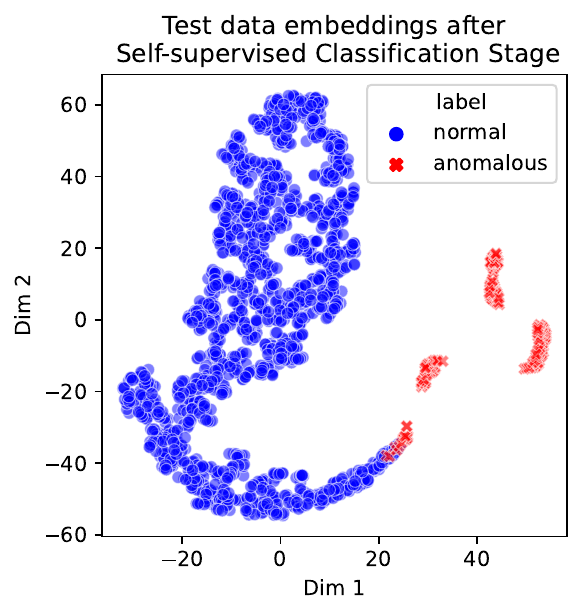}
        \caption{}
        \label{fig:tsne-class}
    \end{subfigure}
    \begin{subfigure}[b]{0.23\textwidth}
        \centering
        \includegraphics[width=\textwidth]{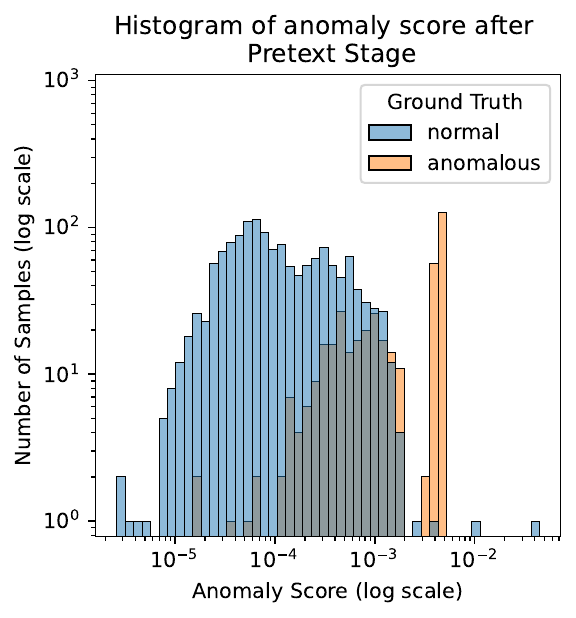}
        \caption{}
        \label{fig:pre-dist}
    \end{subfigure}
    \begin{subfigure}[b]{0.23\textwidth}
        \centering
        \includegraphics[width=\textwidth]{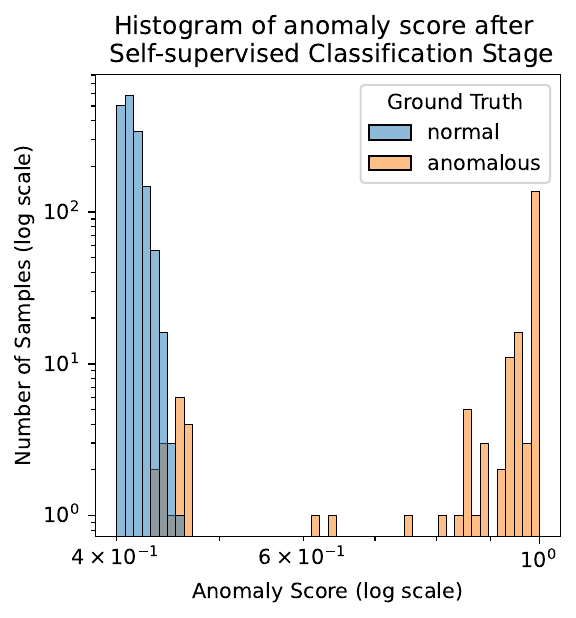}
        \caption{}
        \label{fig:self-prob}
    \end{subfigure}
    \caption{Comparative analysis of the model's stages utilising t-SNE and anomaly score distributions on M-6 dataset in MSL. (\subref{fig:tsne-pretext}) t-SNE after the Pretext stage, indicating the initial representation of normal and anomalous samples. (\subref{fig:tsne-class}) t-SNE after the Self-supervised Classification stage, demonstrating enhanced segregation of normal and anomalous instances. (\subref{fig:pre-dist}) Anomaly score distribution in the Pretext Stage based on Euclidean distances between test and training samples. (\subref{fig:self-prob}) The distribution of anomaly scores following the application of the Self-supervised Classification Stage evidencing an improved representation.}
    \label{fig:stages}
\end{figure*}

In this section, we delve into an in-depth evaluation of the stages of our proposed model, utilising the M-6 time series derived from the MSL dataset. Our primary objective is to develop a comprehensive grasp of the patterns inherent in the features extracted and learned by our model throughout its stages. For this purpose, we employ t-distributed Stochastic Neighbor Embedding (t-SNE) to visualise the output of the Pretext Stage and the Self-supervised Classification stage. These are graphically represented in Figures \ref{fig:tsne-pretext} and \ref{fig:tsne-class} correspondingly.
From these graphical representations, it is palpable that the second stage significantly enhances the discrimination capacity between normal and anomalous samples. It is important to highlight that the first stage reveals some semblance of anomalous samples to normal ones, which might be due to the existence of anomalies close to normal boundaries. However, the Self-supervised Classification stage efficiently counteracts this ambiguity by effectively segregating these instances, thereby simplifying subsequent classification tasks.

Further, we assessed the efficacy of the Self-supervised Classification Stage in refining the representation generated by the Pretext Stage. This was accomplished by juxtaposing the anomaly scores derived from the output of each stage. In Figure~\ref{fig:pre-dist}, the distribution of anomaly scores for the test windows at the Pretext Stage is displayed, computed using the Euclidean distances of the test samples relative to the original time series windows in the training set. Conversely, Figure~\ref{fig:self-prob} depicts the subsequent alteration in the distribution of anomaly scores following the application of the Self-supervised Classification Stage.

In the Pretext Stage, the anomaly score is computed as the minimum distance between the test sample and all original training samples in the representation space, aiming to identify the closest match among the training samples for the given test sample. A smaller distance implies a greater probability that the test sample belongs to the normal data distribution.

In the Self-supervised Classification Stage, we use the inference step in Section \ref{subsec3.2} of the article. Equation \ref{eq:anomalyscore} provides further details on the computation of the anomaly label.

As we can observe from the figures \ref{fig:tsne-pretext} and \ref{fig:tsne-class}, the self-supervised classification Stage has resulted in a significant improvement in the separation of normal and anomalous windows. The distribution of anomaly scores in Figure~\ref{fig:self-prob} is more clearly separated than in Figure~\ref{fig:pre-dist}. This indicates that the second stage has produced more consistent representations. Furthermore, the anomaly scores in Figure~\ref{fig:self-prob} are relatively closer to 0 or 1, indicating improved discrimination between normal and anomalous windows in the second stage of CARLA.

\subsubsection{Positive Pair Selection}
To evaluate the effectiveness of the positive pair selection method in CARLA, we compared two approaches in training: random temporal neighbour from $y$ temporally closest window samples and weak augmentation with noise (add normal noise with sigma 0.01 to a window). We used MSL benchmark and used identical configuration and hyper-parameters for both approaches and evaluated their performance shown in Table \ref{tab:pair}.
\begin{table}[t!]
\centering
\scriptsize
\caption{Positive pair selection results on MSL dataset.}
\label{tab:pair}
\begin{tabular}{c|ccccc}
\hline
\multicolumn{1}{l|}{Pos Pair Selection} & Prec & Rec & F1 & AU-PR \\ \hline
CARLA-noise & 0.3086 & 0.7935 & 0.4433 & 0.3983  \\ 
CARLA-temporal & 0.3891 & 0.7959 & \textbf{0.5227} & \textbf{0.5009} \\\hline
\end{tabular}%
\end{table}
Our experiment demonstrated that selecting positive pairs using a random temporal neighbour is a more effective approach than weak augmentation with noise. Since anomalies occur rarely, choosing a positive pair that is in the target window's temporal proximity is a better strategy. 

\subsubsection{Effectiveness of Different Anomaly Types}
CARLA's performance is evaluated on the MSL dataset by systematically removing different types of anomalies during the Pretext Stage. The evaluation metrics used are precision, recall, F1 score, and AU-PR. The results are sorted based on F1 from the least significant to the most significant types of anomalies in Table \ref{tab:types}. This result represents the overall performance of the CARLA model when using all types of anomalies during the anomaly injection process. It serves as a baseline for comparison with the subsequent results.

\begin{table}[t!]
\centering
\scriptsize
\caption{Effectiveness of different anomaly types on MSL dataset.}
\label{tab:types}
    \begin{tabular}{c|cccc}
    \hline
    Anomaly Type & Prec & Rec & F1 & AU-PR \\
    \hline
    All types & 0.3891 & 0.7959 & \textbf{0.5227}  & \textbf{0.5009} \\
    w/o trend & 0.3185 & 0.7849 & 0.4513 & 0.4972 \\
    w/o contextual & 0.3842 & 0.7935 & 0.5177 & 0.4683 \\
    w/o shapelet & 0.3564 & 0.8370 & 0.4999 & 0.4328 \\
    w/o global & 0.3383 & 0.8404 & 0.4824 & 0.4340 \\
    w/o seasonal & 0.2400 & 0.8428 & 0.3737 & 0.3541 \\
    \hline
    \end{tabular}
\end{table}

In the experiments, anomalies were injected into the training data on a per-window basis. For each window in the multivariate time series, a random number of dimensions was selected, ranging from 1 to $\lceil Dim/10 \rceil $ of the total dimensions. These selected dimensions were injected with anomalies, starting from the same point across all the chosen dimensions. The injected anomaly portion for each dimension varied from 1 data point to 90\% of the window length. This approach ensured a diverse and controlled injection of anomalies within the training data (For more detail, see Algorithm \ref{algo:carlainject})

Eliminating trend anomalies has a significant negative impact on CARLA's precision. However, the recall remains relatively high, indicating that trend anomalies are important for maintaining a higher precision level.
Similar to the previous case, removing contextual anomalies leads to a slight decrease in precision but an improvement in recall.
The exclusion of shapelet anomalies leads to a moderate decrease in precision, while the recall remains relatively high. This suggests that shapelet anomalies contribute to the model's precision but are not as crucial for capturing anomalies in general.
By removing global anomalies, CARLA's precision slightly decreases, indicating that it becomes less accurate in identifying true anomalies. However, the recall improves, suggesting that the model becomes more sensitive in detecting anomalies overall. Removing seasonal anomalies significantly drops precision, though the high recall indicates effective anomaly detection without relying on seasonal patterns.
Overall, the analysis of the results suggests that global and seasonal anomalies play a relatively more significant role in the CARLA model's performance, while contextual, shapelet, and trend anomalies have varying degrees of impact. These findings provide insights into the model's sensitivity to different types of anomalies and can guide further improvements in anomaly injection strategies within CARLA's framework.

\subsubsection{Effectiveness of Loss Components}
To assess the effectiveness of two loss components, namely $\mathcal{L}_{inconsistency}$ and $\mathcal{L}_{entropy}$, in the total loss function for self-supervised classification on the MSL dataset, we conduct experiments using identical architecture and hyper-parameters. The evaluation metrics employed to measure the performance are precision, recall, F1, and AU-PR, as presented in Table \ref{tab:loss}.

\begin{table}[t!]
\centering
\scriptsize
\caption{Effectiveness of $\mathcal{L}_{inconsistency}$ and $\mathcal{L}_{entropy}$ on MSL dataset.}
\label{tab:loss}
\begin{tabular}{cc|ccccc}
\hline
$\mathcal{L}_{inconsistency}$ & $\mathcal{L}_{entropy}$ & Prec & Rec & F1 & AU-PR \\ \hline
\ding{55}  & \ding{55} & 0.3092 & 0.7155 & 0.4318 & 0.3984 \\ 
\ding{55} & \ding{51} & 0.3453 & 0.8112 & 0.4846 & 0.4424 \\ 
\ding{51} & \ding{55} & 0.3219 & 0.7534 & 0.4511 & 0.4175 \\ 
\ding{51} & \ding{51} & 0.3891 & 0.7959 & \textbf{0.5227} & \textbf{0.5009} \\ \hline
\end{tabular}
\end{table}

In cases where only one loss component is used, the model's performance is relatively lower across all metrics, indicating that without incorporating these loss components, the model struggles to detect anomalies in the MSL dataset effectively.

Where both $\mathcal{L}_{inconsistency}$ and $\mathcal{L}_{entropy}$ loss components are included in the total loss function, the model achieves the best performance among all scenarios, surpassing the other combinations. The precision, recall, F1 score, and AU-PR are the highest in this case, indicating that the combination of both loss components significantly improves the anomaly detection capability of the model.

\subsection{Parameter Sensitivity} \label{subsec4.6}
We examine the sensitivity of CARLA's parameters, investigating and analysing the effect of the \textbf{window size}, \textbf{number of classes}, \textbf{number of NN/FN}, and \textbf{entropy coefficient}. 

\subsubsection{Effect of Window Size}
\begin{figure}[t!]
    \centering
    \begin{subfigure}[b]{0.25\textwidth}
        \centering
        \includegraphics[width=\textwidth]{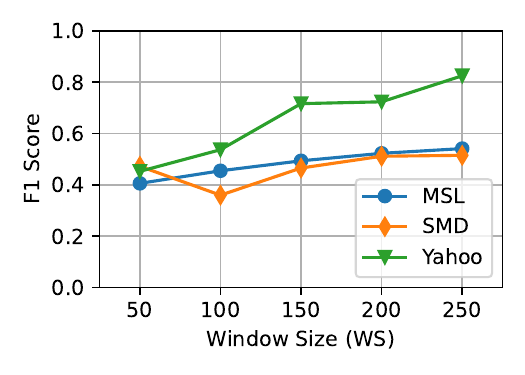}
        \caption{}
        \label{fig:win-f1}
    \end{subfigure}
    \begin{subfigure}[b]{0.25\textwidth}
        \centering
        \includegraphics[width=\textwidth]{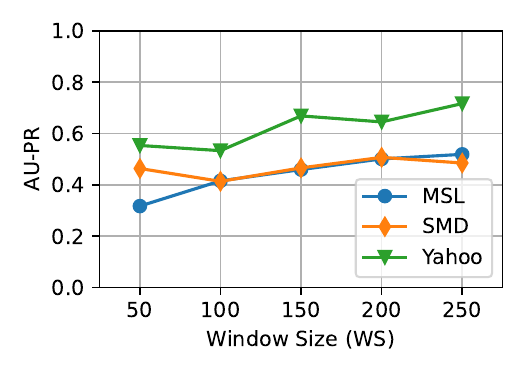}
        \caption{}
        \label{fig:win-aupr}
    \end{subfigure}
    \caption{(a) F1 Score and (b) AU-PR for different window sizes across MSL, SMD, and Yahoo datasets}
    \label{fig:wins}
\end{figure}

As a hyper-parameter in time series analysis, window size holds considerable importance. Table \ref{tab:winsize} and Figure~\ref{fig:wins} 
present the results of exploring the impact of window size on three datasets: MSL, SMD, and Yahoo.
\begin{table*}[t!]
\centering
\scriptsize
\caption{Exploring the effect of window size on MSL, SMD, and Yahoo datasets.}
\label{tab:winsize}
\begin{tabularx}{\textwidth}{c|XXXX|XXXX|XXXX}
\hline
\multicolumn{1}{c}{} & \multicolumn{4}{c}{MSL}          & \multicolumn{4}{c}{SMD}          & \multicolumn{4}{c}{Yahoo}         \\ \hline
WS   & Prec & Rec & F1 & AUPR  & Prec & Rec & F1 & AUPR  & Prec & Rec & F1 & AUPR  \\ \hline
50  & 0.2755 & 0.7702 & 0.4058 & 0.3173 & 0.4569 & 0.4885 & 0.4722 & 0.4636 & 0.3046 & 0.8800 & 0.4526 & 0.5533 \\
100 & 0.3233 & 0.7666 & 0.4548 & 0.4155 & 0.2584 & 0.5938 & 0.3598 & 0.4136 & 0.3709 & 0.9713 & 0.5368 & 0.5331 \\
150 & 0.3606 & 0.7823 & 0.4936 & 0.4595 & 0.3599 & 0.6592 & 0.4655 & 0.4661 & 0.5720 & 0.9578 & 0.7162 & 0.6685 \\
200 & 0.3891 & 0.7959 & 0.5227 & 0.5009 & 0.4276 & 0.6362 & 0.5114 & 0.5070 & 0.5747 & 0.9755 & 0.7233 & 0.6450 \\
250 & 0.4380 & 0.8037 & 0.5412 & 0.5188 & 0.4224 & 0.6596 & 0.5150 & 0.4850 & 0.7129 & 0.9781 & 0.8247 & 0.7164 \\ \hline
\end{tabularx}
\end{table*}

Based on the analysis of both the F1 score and AU-PR, it can be concluded that window size 200 consistently outperforms other window sizes on the MSL, SMD, and Yahoo datasets overall. This window size strikes a balance between precision and recall, effectively capturing anomalies while maintaining a high discrimination ability. Therefore, window size 200 is selected as the best choice in all experiments. 
Additionally, Figure~\ref{fig:wins} shows that our model's performance is stable in window sizes between 150 and 250.  

\subsubsection{Effect of Number of Classes}
\begin{figure}[t]
    \centering
    \begin{subfigure}[b]{0.25\textwidth}
        \centering
        \includegraphics[width=\textwidth]{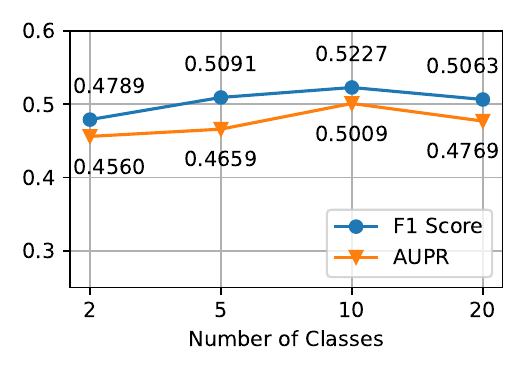}
        \caption{}
        \label{fig:class-f1-aupr}
    \end{subfigure}
    \begin{subfigure}[b]{0.25\textwidth}
        \centering
        \includegraphics[width=\textwidth]{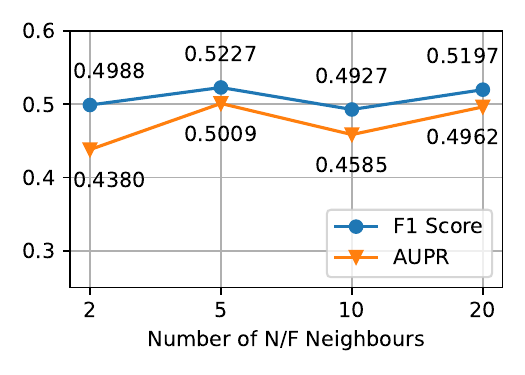}
        \caption{}
        \label{fig:nn-f1-aupr}
    \end{subfigure}
    \begin{subfigure}[b]{0.25\textwidth}
        \centering
        \includegraphics[width=\textwidth]{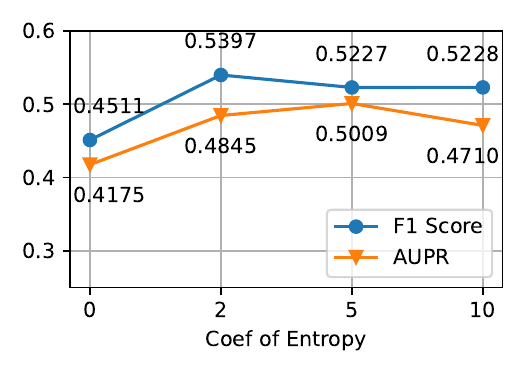}
        \caption{}
        \label{fig:ent-f1-aupr}
    \end{subfigure}
    \caption{F1 Score and AU-PR for (a) different number of classes, (b) different numbers of neighbours in $\mathcal{N}$/$\mathcal{F}$ and (c) different coefficient of entropy, on MSL dataset.}
    \label{fig:class_nn_ent}
\end{figure}

The provided results showcase the performance of the model with varying numbers of classes on the MSL dataset. Based on the illustration of both the F1 score and AU-PR in Figure~\ref{fig:class-f1-aupr}, the model performs relatively well across different numbers of classes. While CARLA's performance is pretty stable across all number of classes denoted on the x-axis of the plot in Figure~\ref{fig:class-f1-aupr}, it can be concluded that using 10 classes yields the highest performance. It strikes a balance between capturing anomalies and minimising false positives while effectively discriminating between anomalies and normal samples. 

The results suggest that increasing the number of classes beyond 10 does not significantly improve the model's performance. When the number of classes increases, normal representations divide into the different classes and, in CARLA, are detected as anomalies (lower probability of belonging to the major class). However, using 2 classes leads to a lower performance compared to 10 classes, implying that a more fine-grained classification with 10 classes provides the assumption that various anomalies are spread in the representation space.

\subsubsection{Effect of Number of neighbours in $\mathcal{N}$ and $\mathcal{F}$}
The provided results in Figure~\ref{fig:nn-f1-aupr} display the evaluation metrics for different numbers of nearest neighbours ($\mathcal{N}$) and furthest neighbours ($\mathcal{F}$) noted as $Q$.

Based on the analysis of both the F1 score and AU-PR, we can see that while CARLA's performance is pretty stable across the different numbers of neighbours in $\mathcal{N}$ and $\mathcal{F}$ denoted on the x-axis of the plot in Figure~\ref{fig:nn-f1-aupr}, it can be concluded that employing 5 nearest/furthest neighbours outperforms the other options. This number of parameters strikes a favourable balance between precision and recall, allowing for effective anomaly detection while maintaining a high discrimination ability. Therefore, selecting 5 neighbours is deemed the optimal choice for this experiment, as it consistently achieves higher F1 scores and AU-PR values.

\subsubsection{Effect of Entropy Coefficient}
In this experiment, we explore the impact of the entropy coefficient. Figure~\ref{fig:ent-f1-aupr} shows the results of evaluating the coefficient of the entropy component in the loss function on the MSL dataset. The coefficients tested are 0, 2, 5, and 10. 

Based on the analysis of both the F1 score and AU-PR, a coefficient value of 5 emerges as the most effective choice for the entropy component in the loss function. It achieves the highest scores for both metrics, indicating better precision-recall balance and accurate ranking of anomalies. However, it is worth noting that coefficients of 2 and 10 also demonstrate competitive performance, slightly lower than the value of 5. The coefficient of 0 significantly reduces the model's performance, further emphasising the importance of incorporating the entropy component for improved anomaly detection performance.

Overall, the analysis indicates that coefficients of 2, 5, and 10 are viable for the entropy component in the loss function, with 5 being optimal due to its superior F1 score and AU-PR.

\section{Conclusion}\label{sec5}
Our innovative end-to-end self-supervised framework, CARLA, utilises contrastive representation learning with anomaly injection to generate robust representations for time series and classify anomalous windows. The use of semantically meaningful nearest and furthest neighbours as a prior allows us to capture underlying patterns in time series and learn a representation that is well-aligned with these patterns, thereby enhancing detection accuracy. Our extensive experimental evaluation of the seven most commonly used time series benchmarks, encompassing 208 datasets with diverse real-world anomalies, shows promising results in detecting anomalies, demonstrating the effectiveness of CARLA in this domain.

This research has demonstrated the potential of contrastive learning combined with synthetic anomaly injection to overcome the limitations of the lack of labelled data in TSAD. To enhance our model's performance, future research will focus on refining positive and negative sample selection strategies. Moreover, we believe that there is much scope to overcome the limitations of the anomaly injection component by broadening the forms of injected anomalies used in our framework. We further hypothesise that varying the severity of injected anomalies may improve the model's ability to detect anomalies in representation space. We commend the investigation of these intriguing prospects in future research.

Additionally, considering that some datasets have a significant number of abnormal samples, resulting in more anomalous windows during the Pretext Stage, it becomes critical to revisit the impact of selecting positives and negatives for abnormal anchors in unsupervised settings. This reconsideration is crucial for enhancing the model's effectiveness in TSAD.

Our results, based on appropriate metrics, contribute significantly to further investigations and advancements in TSAD, particularly on datasets like WADI and KPI. By employing the correct evaluation metrics, such as the conventional F1 score without point adjustment and AU-PR, this research provides a solid foundation for assessing the performance of future models in this field. The findings presented herein enable researchers to benchmark their models and compare their results against established standards. Reliable evaluation metrics and validated results serve as valuable resources, driving innovation in TSAD research.

\appendix

\section{List of Symbols and Definitions} \label{app:symbols}
Table \ref{tab:symbols} demonstrates symbols used in the paper.
\begin{table}[t]
\centering 
\scriptsize
\caption{List of Symbols and Definitions}
\label{tab:symbols}
\begin{tabularx}{0.75\textwidth}{c|X}
\hline
\textbf{Symbol} & \textbf{Definition} \\
\hline
$\mathcal{D}$ & Sequential time series windows where $\mathcal{D}=\{w_1, w_2, .., w_m\}$ \\
$w_i$ & $i^{th}$ window in $\mathcal{D}$ where $w_i=\{x_1, x_2, .., x_{WS}\}$\\
$WS$ & Time series window size\\
$Dim$ & Number of dimensions in time series $\mathcal{D}$\\
$m$ & Number of windows\\
$Q$ & Number of nearest/furthest neighbours \\
$a$ & Anchor sample (i.e. the original time series window) \\
$y$ & Number of temporally proximate windows\\
$p$ & Positive sample (i.e. random window selected from $y$ temporally proximate previous windows)\\
$n$ & Negative sample (i.e. anomaly injected window) \\
$(a, p, n)$ & Triplet in Pretext Stage \\
$\mathcal{T}$ & All triplets set in Pretext Stage \\
$\alpha$ & Margin in pretext loss \\
$\phi_p$ & Pretext trained model \\
$\mathcal{B}$ & All neighbours set \\
$\mathcal{N}$ & $Q$-Nearest neighbours set \\
$\mathcal{F}$ & $Q$-Furthest neighbours set \\
$\phi_s$ & Self-supervised trained model \\
$C_m$ & Majority class label \\
$\beta$ & Entropy loss weight \\
\hline
\end{tabularx}
\end{table}

\setlength{\tabcolsep}{3pt}
\begin{table}[t]
    \centering
    \scriptsize
    \caption{Precision (\(Prec_{PA}\)), recall (\(Rec_{PA}\)), and \(F1_{PA}\) results with point adjustment (PA) for various models on multivariate and univariate time series datasets.}
    \label{tab:PA_results}
    \begin{tabular}{cc}
        % First Subtable
        \begin{subtable}[t]{0.55\textwidth}
            \centering
            \caption{Multivariate Time Series Datasets}
            \label{tab:mts_PA}
            \begin{tabular}{c|c|ccccc}
            \hline
                Model & Metric & MSL & SMAP & SMD & SWaT & WADI \\
                \hline
                \multirow{3}{*}{LSTM-VAE} & $Prec_{PA}$ & 0.8049 & 0.6755 & 0.7765 & 0.9745 & 0.0639 \\
                & $Rec_{PA}$ & 0.8394 & 0.9358 & 0.8740 & 0.7036 & 1.0000 \\
                & $F1_{PA}$ & 0.8218 & 0.7848 & 0.8224 & 0.8172 & 0.1202 \\
                \hline
                \multirow{3}{*}{OmniAnomaly} & $Prec_{PA}$ & 0.8241 & 0.6931 & 0.8193 & 0.9861 & 0.2685 \\
                & $Rec_{PA}$ & 0.9859 & 0.9754 & 0.9531 & 0.7864 & 0.9686 \\
                & $F1_{PA}$ & 0.8978 & 0.8104 & 0.8811 & 0.8750 & 0.4204 \\
                \hline
                \multirow{3}{*}{MTAD-GAT} & $Prec_{PA}$ & 0.4305 & 0.7561 & 0.4106 & 0.9720 & 0.2553 \\
                & $Rec_{PA}$ & 0.9953 & 0.9957 & 0.9370 & 0.6857 & 0.9941 \\
                & $F1_{PA}$ & 0.6011 & 0.8595 & 0.5710 & 0.8052 & 0.4062 \\
                \hline
                \multirow{3}{*}{THOC} & $Prec_{PA}$ & 0.8291 & 0.5741 & 0.3819 & 0.7854 & 0.3871 \\
                & $Rec_{PA}$ & 0.9651 & 0.9324 & 0.8946 & 0.9774 & 0.7638 \\
                & $F1_{PA}$ & 0.8919 & 0.7106 & 0.5353 & 0.8709 & 0.5138 \\
                \hline
                \multirow{3}{*}{AnomalyTran} & $Prec_{PA}$ & 0.7201 & 0.6481 & 0.8058 & 0.9087 & 0.0642 \\
                & $Rec_{PA}$ & 0.9832 & 0.9258 & 0.8482 & 0.7735 & 1.0000 \\
                & $F1_{PA}$ & 0.8313 & 0.7625 & 0.8265 & 0.8357 & 0.1207 \\
                \hline
                \multirow{3}{*}{TranAD} & $Prec_{PA}$ & 0.8911 & 0.8644 & 0.8043 & 0.5207 & 0.0639 \\
                & $Rec_{PA}$ & 0.9953 & 0.9976 & 0.9374 & 0.7779 & 1.0000 \\
                & $F1_{PA}$ & 0.9403 & 0.8906 & 0.9168 & 0.6238 & 0.1202 \\
                \hline
                \multirow{3}{*}{TS2Vec} & $Prec_{PA}$ & 0.2152 & 0.2605 & 0.1800 & 0.1718 & 0.0893 \\
                & $Rec_{PA}$ & 1.0000 & 1.0000 & 0.9950 & 1.0000 & 1.0000 \\
                & $F1_{PA}$ & 0.3542 & 0.4133 & 0.3050 & 0.2932 & 0.1639 \\
                \hline
                \multirow{3}{*}{DCdetector} & $Prec_{PA}$ & 0.4972 & 0.5648 & 0.6359 & 0.6998 & 0.2257 \\
                & $Rec_{PA}$ & 1.0000 & 0.9905 & 0.8110 & 0.9125 & 0.9263 \\
                & $F1_{PA}$ & 0.6642 & 0.7194 & 0.7128 & 0.7921 & 0.3630 \\
                \hline
                \multirow{3}{*}{TimesNet} & $Prec_{PA}$ & 0.7547 & 0.6015 & 0.7962 & 0.8421 & 0.3742 \\
                & $Rec_{PA}$ & 0.9765 & 0.9822 & 0.9265 & 1.000 & 0.9127 \\
                & $F1_{PA}$ & 0.8514 & 0.7461 & 0.8560 & 0.9174 & 0.5308 \\
                \hline
                \multirow{3}{*}{Random} & $Prec_{PA}$ & 0.1746 & 0.9740 & 0.8670 & 0.9398 & 0.8627 \\
                & $Rec_{PA}$ & 1.0000 & 1.0000 & 0.9686 & 0.9331 & 0.9732 \\
                & $F1_{PA}$ & \textbf{0.9840} & \textbf{0.9869} & \textbf{0.9150} & \textbf{0.9364} & \textbf{0.9146} \\
                \hline
                \multirow{3}{*}{CARLA} & $Prec_{PA}$ & 0.8545 & 0.7342 & 0.6757 & 0.9891 & 0.1971 \\
                & $Rec_{PA}$ & 0.9650 & 0.9935 & 0.8465 & 0.7132 & 0.7503 \\
                & $F1_{PA}$ & 0.9064 & 0.8444 & 0.7515 & 0.8288 & 0.3122 \\
                \hline
            \end{tabular}
        \end{subtable}
        &
        % Second Subtable
        \begin{subtable}[t]{0.40\textwidth}
            \centering
            \caption{Univariate Time Series Datasets}
            \label{tab:uts_PA}
            \begin{tabular}{c|c|cc}
            \hline
                Model & Metric & Yahoo-A1 & KPI \\
                \hline
                \multirow{3}{*}{Donut} & \(Prec_{PA}\) & 0.5252 & 0.3928 \\
                & \(Rec_{PA}\) & 0.9280 & 0.5797 \\
                & \(F1_{PA}\) & 0.6883 & 0.4683 \\
             \hline
                \multirow{3}{*}{THOC} & \(Prec_{PA}\) & 0.3138 & 0.3703 \\
                & \(Rec_{PA}\) & 0.9723 & 0.9154 \\
                & \(F1_{PA}\) & 0.4745 & 0.5273 \\
             \hline
                \multirow{3}{*}{TranAD} 
                & \(Prec_{PA}\) & 0.5722 & 0.7012 \\
                & \(Rec_{PA}\) & 0.9858 & 0.7528 \\
                & \(F1_{PA}\) & 0.7241 & 0.7261 \\
             \hline
                \multirow{3}{*}{TS2Vec} & \(Prec_{PA}\) & 0.4993 & 0.4514 \\
                & \(Rec_{PA}\) & 0.9717 & 0.9058 \\
                & \(F1_{PA}\) & 0.6597 & 0.6025 \\
            \hline
                \multirow{3}{*}{DCdetector} & \(Prec_{PA}\) & 0.0750 & 0.2758 \\
                & \(Rec_{PA}\) & 0.7653 & 0.8380 \\
                & \(F1_{PA}\) & 0.1366 & 0.4151 \\
            \hline
                \multirow{3}{*}{TimesNet} 
                & \(Prec_{PA}\) & 0.6232 & 0.5185 \\
                & \(Rec_{PA}\) & 0.8865 & 0.6133 \\
                & \(F1_{PA}\) & 0.7319 & 0.5619 \\
             \hline
                \multirow{3}{*}{Random} & \(Prec_{PA}\) & 0.9020 & 0.6540 \\
                & \(Rec_{PA}\) & 0.9911 & 0.8986 \\
                & \(F1_{PA}\) & \textbf{0.9445} & \textbf{0.7570} \\
             \hline
                \multirow{3}{*}{CARLA} & \(Prec_{PA}\) & 0.7475 & 0.3980 \\
                & \(Rec_{PA}\) & 0.9984 & 0.8933 \\
                & \(F1_{PA}\) & 0.8549 & 0.5507 \\
             \hline
            \end{tabular}
        \end{subtable}
    \end{tabular}
\end{table}
\section{Point Adjustment (PA)} \label{PA-section}
\begin{figure}[t!]
    \centering
    \includegraphics[width=0.5\textwidth]{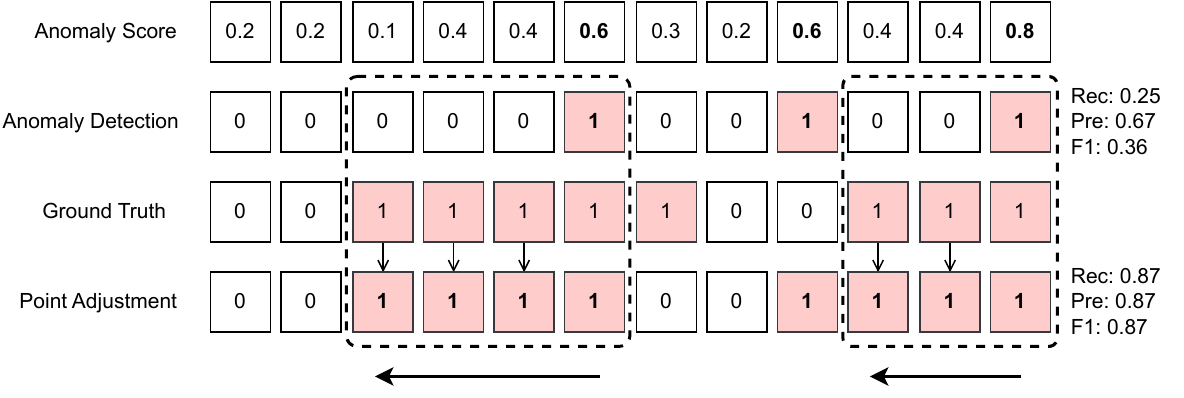}
    \caption{Impact of point adjustment (PA) on anomaly detection in time series. The figure represents a toy example with a sequence of 12 samples, presenting the anomaly scores for each sample. The second row depicts the outcomes of anomaly detection using a threshold of 0.5. The third row provides the actual labels (ground truth). The fourth row presents the adjusted anomaly detection results using point adjustment with the ground truth. Evaluation metrics are calculated for both unadjusted and adjusted anomaly detection.}
    \label{fig:PA}
\end{figure}

A notable issue with many existing TSAD methods is their reliance on the point adjustment (PA) strategy~\cite{xu2018unsupervised}, as used in TSAD literature like~\cite{su2019robust}. PA operates on the principle that if a segment is detected as anomalous, all previous segments in that anomaly subsequence are also considered anomalies (see Figure~\ref{fig:PA}). Incorporating PA, the Random model outperform all state-of-the-art models, as depicted in Tables \ref{tab:mts_PA} and \ref{tab:uts_PA}. 

\newpage
\bibliographystyle{elsarticle-harv}

\bibliography{cas-refs}

\end{document}